\documentclass{sig-alternate}

\usepackage[usenames,dvipsnames]{xcolor}

\usepackage{amsfonts}
\usepackage{amssymb}
\usepackage{amsmath}
\usepackage{latexsym}
\usepackage{subfigure}
\usepackage{verbatim}
\usepackage{url}
\usepackage{color}
\usepackage{array}
\usepackage{textcomp}
\usepackage{paralist}

\usepackage{hyperref}
\usepackage{url}
\usepackage{xspace}

\usepackage[usenames,dvipsnames]{xcolor}

\usepackage{times}

\newcommand{\eat}[1]{}

\eat{
\newcommand{\mikef}[1]{}
\newcommand{\srm}[1]{}
\newcommand{\barzan}[1]{}
\newcommand{\mikej}[1]{}
\newcommand{\purna}[1]{}
}

\newcommand{\un}[0]{Uncertainty\xspace}
\newcommand{\ete}[0]{MinExpError\xspace}
\newcommand{\smd}[0]{MarginDistance\xspace}
\newcommand{\pst}[0]{Bootstrap-LV\xspace}
\newcommand{\pba}[0]{PBA\xspace}
\newcommand{\crowdER}[0]{CrowdER\xspace}
\newcommand{\iwal}[0]{IWAL\xspace}
\newcommand{\para}[0]{ParaActive\xspace}
\newcommand{\aditya}[0]{CVHull\xspace}

\newcommand{\ent}[0]{Entropy\xspace}

\eat{

}

\makeatletter
\ifnum\inputlineno=\m@ne
\let\showlineno\@empty
\else
\def\showlineno{ line \the\inputlineno}
\fi
\makeatother

\begin{document}

\title{Active Learning for Crowd-Sourced Databases\titlenote{A shorter version of this manuscript has been published in 
Proceedings of Very Large Data Bases \cite{mozafari_pvldb2015_active}.}}

\numberofauthors{5} 



\author{
Barzan Mozafari\\
       \affaddr{University of Michigan, Ann Arbor}\\
       \email{mozafari@umich.edu}
\alignauthor
Purna Sarkar\\
       \affaddr{UC Berkeley}\\
       \email{psarkar@cs.berkeley.edu}
\and
\alignauthor
Michael Franklin\\
       \affaddr{UC Berkeley}\\
       \email{franklin@cs.berkeley.edu}
\alignauthor 
Michael Jordan\\
      \affaddr{UC Berkeley}\\
       \email{jordan@cs.berkeley.edu}
\alignauthor ~~~~~~~Samuel Madden\\
       \affaddr{MIT}\\
       \email{madden@csail.mit.edu}
}


\maketitle


\begin{abstract}
Crowd-sourcing has become a popular means of acquiring labeled data for many tasks where humans are more accurate than computers, such as image tagging, entity resolution, or sentiment analysis. However, due to the time and cost of human labor, solutions that solely rely on crowd-sourcing are often limited to small datasets (i.e., a few thousand items). This paper proposes algorithms for integrating machine learning into crowd-sourced databases in order to combine the accuracy of human labeling with the speed and cost-effectiveness of machine learning classifiers. By using \emph{active learning} as our optimization strategy for labeling tasks in crowd-sourced databases, 
we can minimize the number of questions asked to the crowd, allowing crowd-sourced applications to \emph{scale} (i.e, label much larger datasets at lower costs).

Designing active learning algorithms for a crowd-sourced database poses many practical challenges: such algorithms need to be generic, scalable, and easy-to-use for a broad range of practitioners, even those who are not machine learning experts. We draw on the theory of nonparametric bootstrap to design, to the best of our knowledge, the first active learning algorithms that meet all these requirements.

Our results, on $3$ real-world datasets collected with Amazon's Mechanical Turk, and on $15$  UCI datasets, show that our methods on average ask $1$--$2$ orders of magnitude fewer questions than the baseline, and $4.5$--$44\times$ fewer than existing active learning algorithms. 

\comment{
  Crowd-sourcing has become a popular means of acquiring labeled data
  for  a variety of 
  tasks \tofix{where humans are more accurate than
  computers}, such as image tagging, entity resolution, or sentiment analysis. 
  However,
   solutions that solely rely on the crowd to provide data  are often
  limited to small datasets (i.e., a few thousand items), 
    due to the time and monetary cost of asking humans. 
  In this paper, we propose algorithms for integrating machine
  learning into crowd-sourced databases, particularly for labeling tasks,
where our goal is to allow
  crowd-sourcing applications to  \emph{scale}, i.e., handle larger
  datasets at lower costs.  
The key observation is that humans and machine learning classifiers can be
  complementary in many of the tasks above: humans are often more accurate but slow and
  expensive, while classifiers are less accurate, but faster
  and cheaper. 
Based on this observation, we propose 
using {\it active learning}  
as the optimization strategy for labeling tasks  in crowd-sourced databases, whereby  humans and classifiers are combined to minimize the number of questions asked to the crowd.

Designing active learning algorithms that can be deployed in a crowd-sourced database
poses many practical challenges:
such algorithms need to be generic, scalable, and easy-to-use in order
to be useful to a broad range of practitioners who are not necessarily 
machine learning experts.
We draw on the theory of nonparametric bootstrap to design, to the best of our knowledge,
the first active learning algorithms that meet all these requirements.

 Our results, on $3$ \tofix{real-world} datasets
  collected with Amazon's Mechanical Turk, and on 15 well-known UCI
  datasets, 
  show that our methods on average ask $1$--$2$ orders of
  magnitude fewer questions than the baseline, and  $4.5$--$44\times$ fewer questions than the state-of-the-art active learning algorithms.
 } 
  
\comment{
, making them
 applicable to almost arbitrary classifiers,
 (ii) are amenable to embarrassingly parallel execution,
 (iii) treat the classifier as a block-box, with no modifications to 
the user-code---an important requirement in many scientific and industrial settings 
that have developed  
sophisticated classifiers over the years, and finally, 
(iv) automatically adjust the required degree of redundancy 
for each individual label to minimize the human-noise. 
}

\end{abstract}


\vspace*{-0.1in}
\section{Introduction}
\label{sec:intro}

\begin{figure}[t]
\vspace{0.1in}
\centering\includegraphics[width=3.4in]{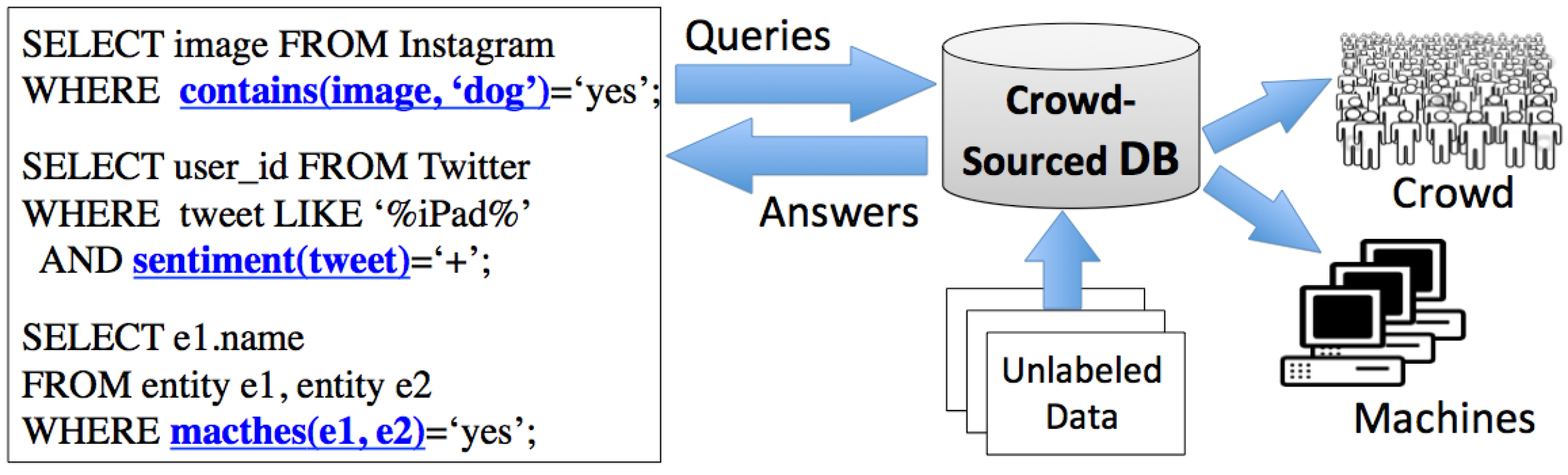}
\caption{{\small Examples of Labeling Queries in a Crowd-sourced DB.}}
\label{fig:arch}
\end{figure}

Crowd-sourcing marketplaces, such as Amazon's Mechanical Turk, have made it
easy to recruit a crowd of people to perform tasks that are difficult for
computers, such as
 entity resolution~\cite{arasu-er,active-sampling,Qurk,sarawagi-er,crowd-er}, 
image annotation~\cite{vision-al}, and sentiment analysis~\cite{sentiment-twitter}.
Many of these tasks can be modeled as database problems, where each item is represented as a row with some missing attributes
(labels) that the crowd workers supply. 
This has given rise to a new generation of database systems, called
\emph{crowd-sourced databases}~\cite{crowdDB, CrowdForge, Qurk, deco}, that enable users to issue more powerful queries 
 by combining human-intensive tasks with traditional query processing techniques. Figure~\ref{fig:arch} provides a few examples 
 of such queries, where part of the query is processed by the machine (e.g., whether the word ``iPad'' 
 appears in a tweet)
 while the \emph{human-intensive} part
  is sent to the crowd for labeling (e.g., to decide if the tweet has a positive sentiment).

While query optimization techniques~\cite{crowdDB, crowd-counting, crowd-filter, deco} can reduce the number of items that need 
to be labeled,
any crowd-sourced database that solely relies on human-provided labels will eventually 
suffer from scalability issues when faced with web-scale datasets and problems (e.g., daily tweets or images).
This is because
 labeling each item by humans can cost several cents and take several minutes.
For instance, given the example of Figure~\ref{fig:arch}, 
even if we filter out tweets that do not contain ``iPad'', 
there could still be millions of tweets with ``iPad'' that require sentiment labels (`positive' or  `non-positive').

To enable  crowd-sourced databases to scale up to large datasets, 
 we advocate combining humans and machine learning algorithms (e.g., classifiers), 
 where (i) the crowd labels items that are either 
 inherently \emph{difficult} for the algorithm, or if labeled, will form the \emph{best training data} for the algorithm, and 
  (ii) the (trained) algorithm is used to label the remaining items much more quickly and
cheaply.
In this paper, we focus on {\it labeling} algorithms (classifiers) that assign one of several discrete values to each item, as opposed to predicting numeric values (i.e., regression), or 
 finding missing items~\cite{getting-it-all}, leaving these other settings for future work.

  Specifically, given a large, unlabeled dataset (say, millions of images) and a classifier that 
can attach a label to each unlabeled item (after sufficient training), 
our goal is 
to determine \emph{which} questions to ask the crowd in order to
(1)  achieve the best training data and overall accuracy, given time or budget constraints, or (2)  minimize the number of questions, given a desired level of accuracy. 

  Our problem is closely related to the  classical problem of {\it active
  learning (AL)}, where the objective is to select
statistically optimal training data~\cite{jordan-96}.
However, in order for an AL algorithm to be a 
practical 
optimization strategy for labeling tasks in a
crowd-sourced database, it must 
satisfy a number of {\it systems} challenges and criteria that have not
been a focal concern in traditional AL literature, as described next. 


\subsection{Design Criteria}
An AL algorithm must meet the following criteria to 
 be used as the default optimization strategy in a crowd-sourced database:

\begin{asparaitem}

\vspace{0.1in}
\item[] \textbf{1. Generality.}
Our system must come with a few built-in AL algorithms that are applicable to arbitrary classification and labeling tasks, as crowd-sourced systems are used in a wide range of different domains. In Figure~\ref{fig:arch}, for example, one query involves sentiment analysis while another seeks images containing a dog. 
Clearly, these tasks require drastically different classifiers.  
Although our system allows expert users to provide their own custom-designed AL that works for their classification algorithm, most users may only have a classifier. Thus, to support general queries, our AL algorithm should make minimal or no assumptions about the classifiers that users provide with their labeling tasks. 


\vspace{0.1in}
\item[] \textbf{2. Black-box treatment of the classifier}.
Many AL algorithms that provide theoretical guarantees,  
need to \emph{access} and \emph{modify} the internal logic of the given classifier 
 (e.g., adding constraints to the classifier's internal loss-minimization step~\cite{iwal}).
Such modifications are acceptable in theoretical settings but not in real-world applications, as state-of-the-art classifiers used in science
and industry are rarely a straightforward implementation of textbook
algorithms.
Rather, these finely-tuned classifiers typically use thousands of lines of code to implement
many intertwined
 steps (e.g., data cleaning, feature selection, parameter tuning, heuristics, etc.). 
 In some
cases, moreover, these codebases use proprietary libraries that cannot be modified. 
 Thus, to make our crowd-sourcing
system useful to a wide range of practitioners and scientists
(who are not necessarily machine learning experts), we need an AL
algorithm that treats the classifier as a black-box, i.e., no
 modifications to the internals of the classifier.

\vspace{0.1in}
\item[] \textbf{3. Batching.}
Many (theoretical) AL algorithms are designed for online (a.k.a. streaming) scenarios in which items are revealed \emph{one at a time}. 
This means that the AL algorithm decides whether to request a label for the current item, and if so, 
awaits the label before proceeding to the next item.
While these settings are attractive for theoretical 
reasons, 
they are unrealistic in practice. First, we often have access 
to a large pool of unlabeled data to choose from (not just the next item), which should allow us to make better choices.
Second, online AL settings typically perform an (expensive) analysis for each item~\cite{iwal, iwalbb, general-agnostic}, rendering them computationally prohibitive.
Thus, for efficiency and practicality, the AL algorithm must support 
\emph{batching},\footnote{Batch support  is challenging 
because
 AL usually involves case-analysis for different combinations of labels and items. These combinations 
grow exponentially in the number of items in the batch, \emph{unless} most of the analysis can be shared 
among different cases.} so that (i) the analysis is done only once per each \emph{batch of multiple items},
and (ii) items are sent to the crowd in batches (to be labeled in parallel).

\vspace{0.1in}
\item[] \textbf{4. Parallelism.}
We aim to achieve \emph{human-scalability} (i.e., asking the crowd fewer questions) through AL.
However, we are  also concerned with \emph{machine-scalability},
 because AL often involves repeatedly training a 
 classifier and can thus be computationally expensive.
 While  AL has been historically applied to small datasets,
increasingly massive datasets (such as those motivating this paper) pose new computational challenges. 
Thus, a design criterion in our system is that our AL algorithm must be  amenable to  parallel execution on modern many-core processors and distributed clusters.

\vspace{0.1in}
\item[] \textbf{5. Noise management.} AL has traditionally dealt with expert-provided labels that are often taken as 
ground truth (notable exceptions are agnostic AL approaches~\cite{general-agnostic}). In contrast, crowd-provided labels are subject to a greater degree of noise, e.g., innocent errors, typos, lack of domain knowledge, and even deliberate spamming.

\end{asparaitem}


AL has a rich literature in machine learning~\cite{al-survey}. 
However, the focus has been largely theoretical, with concepts from 
learning theory used to establish bounds on sample complexity, but leaving
a significant gap between theory and practice.  Rather than aiming at
such theoretical bounds, this paper focuses on 
a set of practical design criteria
and provides sound heuristics for the AL problem; the origin of these criteria
is in real-world and systems considerations (in particular, issues of scale and ease-of-use).

To the best of our knowledge, no existing AL algorithm 
 satisfies \emph{all} of the aforementioned requirements. 
 For example,  those AL algorithms that are general~\cite{iwal,iwalbb,general-agnostic} do not support batching or parallelism and often require modifications to the classifier.
 (See Section \ref{sec:related} for a detailed literature review.)
In this paper, we design the first AL algorithms that meet \emph{all} these design criteria,  paving 
the way towards a scalable and generic crowd-sourcing system that can be 
 used by a wide range of practitioners.

\subsection{Our Contributions}


Our main contributions are two 
AL algorithms, called {\it \ete} and {\it \un}, along with a  noise-management technique, called {\it partitioning-based allocation (PBA)}.
The \un algorithm requests labels for the  items that the  classifier is \emph{most uncertain} about. 
We also design a more sophisticated algorithm, called \ete, that combines the current quality (say, accuracy) of the classifier with its uncertainty in a mathematically sound way, in order to choose the best questions to ask.
\un is faster than \ete, but it also has lower overall accuracy, especially in the \emph{upfront} scenario, i.e., where we 
 request all labels in a single batch.
 We also study  the \emph{iterative} scenario, i.e., where we request labels in multiple batches and refine our decisions after 
 receiving each batch.


A major novelty of our AL algorithms is in their use of bootstrap theory, \footnote{
A short background on bootstrap theory is provided in Section~\ref{sec:bootstrap}.} which yields several key advantages.
 First, bootstrap 
can deliver consistent estimates for a large class of estimators,\footnote{Formally, this class includes any function that is Hadamard differentiable, 
including  $M$-estimators (which include most machine learning algorithms and maximum likelihood estimators~\cite{M_estim_bootstrap}).
}  making our AL algorithms general and applicable to nearly any classification task. 
Second, bootstrap-based estimates 
can be obtained while treating the classifier as a complete black-box. 
Finally, the required bootstrap computations  can be performed independently from each other, hence,
 allowing for an embarrassingly parallel execution.

Once \ete or \un decides which items to send to the crowd, 
dealing with the inherent noise in crowd-provided labels is the next challenge.
A common practice is to use \emph{redundancy}, i.e., to ask each question to multiple workers.
However, instead of applying the same degree of redundancy to all items, we have developed a novel technique based on integer linear programming,
 called \pba, which dynamically partitions the unlabeled items based on their degree of difficulty for the 
 crowd
and determines the required degree of redundancy for each partition.

Thus, with a careful use of bootstrap theory as well as our batching and noise-management techniques, 
our AL algorithms meet \emph{all} our requirements for building a practical crowd-sourced system, namely 
generality, scalability (batching and parallelism), and ease-of-use (black-box view and automatic noise-management).

 We have evaluated the effectiveness of our
 algorithms on $18$ real-world datasets ($3$ crowd-sourced using Amazon Mechanical Turk, and $15$
well-known datasets from the UCI KDD repository). Experiments show that our AL algorithms achieve the same quality
as several existing approaches while
 significantly reducing the number of questions asked of the crowd. 
 Specifically, on average, we reduce the number of  questions asked by:
\begin{itemize}
\item $100\times$ ($7\times$) in the upfront (iterative) scenario, compared to passive learning, and
\item $44\times$ ($4.5\times$) in the upfront (iterative) scenario, compared to \iwal~\cite{para-active, iwal, iwalbb}
which is the state-of-the-art general-purpose AL algorithm.
\end{itemize}
    Interestingly,  we also find that
our algorithms (which are general-purpose) are still
 competitive with, and sometimes even superior to,  some of the
 state-of-the-art domain-specific (i.e., less general) AL techniques.
 For example, our algorithms ask:
 \begin{itemize}
 \item $7\times$  fewer questions than 
\crowdER~\cite{crowd-er}  and  an order of magnitude fewer than \aditya~\cite{active-sampling},
which are among the most recent AL algorithms for entity resolution in database literature,
\item $2$--$8\times$ fewer questions than \pst~\cite{provost} and \smd~\cite{svm-margin}, and
\item  $5$--$7\times$ fewer questions for SVM classifiers than  AL techniques that are specifically designed for SVM~\cite{svm-margin}.
\end{itemize}

\vspace{-0.1in}
\section{Overview of Approach}
\label{sec:background}

Our approach in this paper is as follows. The user provides
 (i) a pool of unlabeled items (possibly with some labeled
items as initial training data), (ii) a classifier (or ``learner'')
that improves with more and better training data, and a specification
as to whether learning should be {\it upfront} or {\it iterative}, and
(iii) a budget or goal, in terms of time, quality, or cost (and a
pricing scheme for paying the crowd).

Our system can operate in two different scenarios,  {\it upfront} or {\it iterative},  
 to train the classifier and label the items (Section~\ref{sec:regimes}).
  Our proposed AL algorithms, \un{}
and \ete{}, can be used in either scenario (Section~\ref{sec:approach}).
Acquiring labels from a crowd  raises 
interesting issues, such as how  best to employ redundancy to
minimize error, and how many questions to asked the crowd
 (Sections \ref{sec:noise} and \ref{sec:optimization}).
Finally, we empirically evaluate our  algorithms (Section~\ref{sec:expr}).

\subsection{Active Learning Notation}
\label{sec:statement}

An active learning algorithm is typically composed of (i) a \emph{ranker}
$\mathcal{R}$, (ii) a \emph{selection strategy} $\mathcal{S}$, 
 and (iii) a budget allocation strategy $\Gamma$. The ranker $\mathcal{R}$ takes as input
a classification algorithm\footnote{For ease of presentation, in this paper we assume binary classification (i.e., $\{0,1\}$), but our work  applies to arbitrary classifiers.} $\theta$,
a set of labeled items $L$, and
a set of unlabeled items $U$, and 
 returns as output an ``effectiveness'' score $w_i$ for each unlabeled item $u_i\in U$.
Our proposed algorithms in Section~\ref{sec:approach} are essentially ranking algorithms that produce these scores.
A selection strategy then uses 
the scores returned from the ranker 
to choose a subset $U'\subseteq U$ which will be sent for human labeling.
For instance, one selection strategy is picking the top $k$ items with the largest (or smallest) scores,
where $k$ is determined by the budget or quality requirements. 
In this paper, we use weighted sampling to choose $k$ unlabeled items, where
 the probability of choosing each item is proportional to its score.
Finally, once $U'$ is chosen, a budget allocation strategy $\Gamma$ 
decides how to best acquire labels for all the items in $U'$:
 $\Gamma(U',B)$ for finding the
 most accurate labels given a fixed budget $B$,
 or  $\Gamma(U',Q)$ for the cheapest labels given a minimum quality
requirement $Q$.
For instance, to reduce the crowd noise, a common strategy is to ask each question to multiple labelers and take the majority vote.
 In Section \ref{sec:noise}, we  introduce our {\it Partitioning Based Allocation (\pba)} algorithm, which will be
our choice of $\Gamma$ in this paper. 
 
\subsection{Active Learning Scenarios}
\label{sec:regimes}
 
This section describes how learning works in the {\it upfront} and the {\it iterative} scenarios. 
Suppose we have a given budget $B$ for asking questions (e.g., in terms of money, time, or number of questions)
or a quality requirement $Q$ (e.g., required accuracy or F1-measure\footnote{F1-measure is the harmonic mean of precision and recall and is frequently used to assess the quality of a classifier.  
})
that our classifier must achieve.

The choice of the scenario depends on
user's preference and needs.
 Figure~\ref{fig:upfront} is the pseudocode of the {\it upfront} scenario.
In this scenario, 
the ranker 
computes effectiveness scores solely based on
 the initial labeled data\footnote{In the absence of initially labeled data, we can first spend part of the budget to
label a small, random sample of the data.}, $L_0$. 
Then, a subset $U'\subseteq U$ is chosen and sent to the crowd (based on $\mathcal{S}$ and, $B$ or $Q$). While waiting for the crowd to label $U'$ (based on $\Gamma$ and, $B$ or $Q$), 
we train our classifier $\theta$ on $L_0$ to label the remaining items, namely $U-U'$, and 
immediately send their labels back to the user. 
Once crowd-provided labels arrive, they are also sent to the user.
Thus, the final result consists of the union of these two labeled sets.

 Figure~\ref{fig:iterative} shows the pseudocode of the {\it iterative} scenario.
 In this scenario, we ask for labels in several iterations. 
We ask the crowd to label a few items, adding those labels to the existing training set, and retrain.
Then, we choose a new set of unlabeled items and iterate until 
we have exhausted our budget $B$ or met our quality goal $Q$.
At each iteration,  our allocation strategy ($\Gamma$)  seeks the cheapest or most accurate labels for 
the chosen items ($U'$),
then
our ranker uses the original training data $L_0$ as well as  
 the crowd labels $CL$ collected thus far  to decide how to score the remaining unlabeled items.


 Note that the upfront scenario is \emph{not} an iterative scenario with a single iteration, because
 the former does not use the crowd-sourced labels in training the classifier.
 This difference is important as different applications
may call for different scenarios. 
When early answers are strictly preferred, say in an interactive
search interface, the upfront scenario can immediately feed users
 with model-provided labels 
until the crowd's answers arrive for the remaining items.
The upfront scenario is also preferred when 
the project has a stringent accuracy requirement that 
only  \emph{gold}  data 
(here, $L_0$)  be used for training the classifier, say to avoid the potential noise of crowd  labels in the training phase.
In contrast, the iterative scenario is 
computationally slower, as it has to repeatedly retrain a classifier and
wait for crowd-sourced labels. However, it can adaptively adjust its
scores in each iteration, thus achieving a smaller error for the same budget than the
upfront one. This is because the upfront scenario
must choose all the items it wants labeled at once, based only on
a limited set of initial labels.

{
\begin{figure}[t]
\small
\begin{tabbing}
\ \ \ \ \ \= \ \ \ \ \= \ \ \ \ \= \ \ \ \ \= \ \ \ \ \= \ \ \ \ \=\\
\underline{\textbf{Upfront Active Learning ($B$ or $Q$, $L_0, U, \theta, \mathcal{R}, \mathcal{S}, \Gamma$)}}\\
\textbf{Input:} $B$ is the total budget (money, time, or number of questions),\\
\>\>~~~$Q$ is the quality requirement (e.g., minimum accuracy, F1-measure),\\
\>\>~~~$L_0$ is the initial labeled data,\\
\>\>~~~$U$ is the unlabeled data,\\
\>\>~~~$\theta$ is a classification algorithm to (imperfectly) label the data,\\
\>\>~~~$\mathcal{R}$ is a ranker that gives ``effectiveness'' scores to unlabeled items,\\
\>\>~~~$\mathcal{S}$ is a selection strategy (specifying which items should be labeled  \\
\>\>~~~~~~~~~by the crowd,  given their effectiveness scores).\\
\>\>~~~$\Gamma$ is a budget allocation strategy for acquiring labels from the crowd\\
\textbf{Output:} $L$ is the labeled version of $U$\\
\textbf{1:}\>$W \leftarrow \mathcal{R}(\theta, L_0, U)$ ~~~\textbf{//$w_i\in W$ is the effectiveness score for $u_i\in U$}\\
\textbf{2:}\>Choose $U'\subseteq U$ based on $\mathcal{S}(U,W)$ such that $U'$ can be labeled with\\
\textbf{3:}\>\>\>budget $B$ ~or~ $\theta^{L_0}(U-U')$ satisfies $Q$\\
\textbf{4:}\>$M$$L$$\leftarrow \theta^{L_0}(U - U')$ ~~~\textbf{//train $\theta$ on $L_0$ to automatically label $U - U'$}\\
\textbf{5:}\>Immediately display $ML$ to the user ~~~\textbf{//Early query results}\\
\textbf{6:}\>$CL\leftarrow \Gamma(U', B$ \textbf{or} $Q)$ ~~~\textbf{//Ask (and wait for) the crowd to label $U'$}\\
\textbf{7:}\>$L\leftarrow CL \cup ML$ ~~~\textbf{//combine crowd and machine provided labels}\\
\underline{\textbf{Return $L$}~~~~~~~~~~~~~~~~~~~~~~~~~~~~~~~~~~~~~~~~~~~~~~~~~~~~~~~~~~~~~~~~~~~~~~~~~~~~~~~~~~~~~~~~~~~~~}
\end{tabbing}
\vspace{-.2in}
\caption{The upfront scenario in active learning.\vspace*{-2.8cm}}
\label{fig:upfront}
\end{figure}
}

{
\begin{figure}[t]
\small
\begin{tabbing}
\ \ \ \ \ \= \ \ \ \ \= \ \ \ \ \= \ \ \ \ \= \ \ \ \ \= \ \ \ \ \=\\
\underline{\textbf{Iterative Active Learning ($B$ or $Q$, $L_0,U, \theta, \mathcal{R}, \mathcal{S}, \Gamma$)}}\\
\textbf{Input:} Same as those in Figure~\ref{fig:upfront})\\
\textbf{Output:} $L$ is the labeled version of $U$\\
\textbf{1:}\>$CL\leftarrow\emptyset$ ~~~// \textbf{labeled data acquired from the crowd}\\
\textbf{2:}\>$L\leftarrow \theta^{L_0}(U)$ ~~~\textbf{//train  $\theta$ on $L_0$ \& invoke it to label $U$}\\
\textbf{3:}\>While our budget $B$ is not exhausted or $L$'s quality does not meet $Q$:\\
\textbf{4:}\>\>$W \leftarrow \mathcal{R}(\theta, L_0\cup CL, U)$ ~~~\textbf{//$w_i\in W$ is the effective score for $u_i\in U$}\\
\textbf{5:}\>\>Choose $U'\subseteq U$ based on $\mathcal{S}(U, W)$~~ (subject to $B$ or $Q$)\\
\textbf{6:}\>\>$L'\leftarrow \Gamma(U', B$ \textbf{or} $Q)$  ~~~\textbf{//Ask (and wait for) the crowd to label $U'$}\\
\textbf{7:}\>\>$CL\leftarrow CL\cup L'$, ~~$U \leftarrow U - U'$~\textbf{//remove crowd labels from $U$}\\
\textbf{8:}\>\>$L \leftarrow CL\cup \theta^{L_0\cup CL}(U)$~~~\textbf{//train  $\theta$ on $L_0\cup CL$ to label remaining  $U$}\\
\underline{\textbf{Return $L$}~~~~~~~~~~~~~~~~~~~~~~~~~~~~~~~~~~~~~~~~~~~~~~~~~~~~~~~~~~~~~~~~~~~~~~~~~~~~~~~~~~~~~~~~~~~~~}
\end{tabbing}
\vspace{-.2in}
\caption{The iterative scenario in active learning.}
\label{fig:iterative}
\end{figure}
}

\section{Ranking Algorithms}
\label{sec:approach}

\newcommand{\var}{\mbox{Var}}

This paper 
 proposes two novel AL algorithms, \un and \ete.
AL algorithms consist of (i) a ranker $\mathcal{R}$ that assigns scores to  unlabeled items, 
 (ii) a selection 
 strategy  $\mathcal{S}$ that uses these scores to choose which items to label, 
and (iii) a budget allocation strategy $\Gamma$ to decide \emph{how} to acquire crowd labels for those chosen items. 
As explained in Section~\ref{sec:statement}, our AL algorithms use 
 weighted sampling and \pba (introduced in Section \ref{sec:noise})
as their selection and budget allocation strategies, respectively.
Thus, for simplicity, we use \un and \ete to refer to both our AL algorithms 
and their corresponding rankers. 
Both rankers can be used in either \emph{upfront} or \emph{iterative}
scenarios.
Section~\ref{sec:bootstrap} provides brief background on nonparametric bootstrap theory,
which we use in our rankers. Our rankers are introduced in Sections \ref{sec:uncertainty} and \ref{sec:ete}.


\subsection{Background: Nonparametric Bootstrap}
\label{sec:bootstrap}

Our ranking algorithms rely on \emph{nonparametric bootstrap} (or simply the \emph{bootstrap}) to assess the benefit of acquiring labels for different unlabeled 
items. 
 Bootstrap~\cite{efron93bootstrap} is a powerful statistical technique traditionally designed for estimating the uncertainty of estimators.   
Consider an estimator $\theta$ (say, a classifier) that can be learned from data $L$ (say, some training data) to
estimate some value of interest for a data point $u$ (say, the class label of $u$). This estimated value, denoted as $\theta^L(u)$,
is a point-estimate (i.e., a single value), and hence, reveals little information about how this value would change
if we used a different data $L'$. This information is critical as most real-world datasets are noisy, subject-to-change, or
 incomplete. For example, in our active learning context, 
missing data means that we can only access part of
our training data. Thus, $L$ should be treated a random variable drawn from some (unknown)
 underlying distribution $\mathbb{D}$.
 Consequently, statisticians are often interested in measuring \emph{distributional information} about $\theta^L(u)$, 
 such as variance,
 bias, etc.  Ideally, one could measure such statistics by (i) drawing many new datasets, say $L_1,\cdots,L_k$ for some large $k$, from the same  distribution $\mathbb{D}$ that generated  the original $L$,\footnote{A common assumption in nonparametric bootstrap is that
 $L$ and $L_i$ datasets are independently and identically drawn (I.I.D.) from $\mathbb{D}$.}
 (ii) computing $\theta^{L_1}(u), \cdots, \theta^{L_k}(u)$, 
 and finally (iii) inducing a distribution for $\theta(u)$ based on the observed values of $\theta^{L_i}(u)$.
We call this true distribution $D^\theta(u)$ or simply $D(u)$ when $\theta$ is understood.
Figure~\ref{fig:ideal} illustrates this computation. 
For example, when $\theta$ is a binary classifier, $D(u)$ is simply a histogram with two bins ($0$ and $1$), where
the value of the $i$'th bin for $i\in\{0,1\}$ is $Pr[\theta^L(u)=i]$ when $L$ is drawn from $\mathbb{D}$.
Given $D(u)$, any distributional information (e.g.,  variance) can be obtained. 

Unfortunately, in practice   
the underlying distribution $\mathbb{D}$ is often unknown, 
and hence, direct computation of $D(u)$ using the procedure of Figure \ref{fig:ideal}  is impossible. 
This is where bootstrap~\cite{efron93bootstrap} becomes useful.
The main idea of bootstrap is simple: treat $L$ as a proxy for its underlying distribution $\mathbb{D}$. 
In other words, instead of drawing $L_i$'s directly from $\mathbb{D}$, generate new datasets $S_1, \cdots, S_k$ by resampling
from $L$ itself. Each $S_i$ is called a (bootstrap) replicate or simply a bootstrap.
 Each $S_i$ 
is generated by drawing $n=|L|$ I.I.D. samples 
\emph{with replacement} from $L$, and hence, some elements of $L$ might be repeated or missing in $S_i$. 
Note that all bootstraps have the same cardinality as $L$, i.e. $|S_i|=|L|$ for all $i$.
By computing $\theta$ on these bootstraps, namely $\theta^{S_1}(u),\cdots,\theta^{S_k}(u)$, we can create an empirical distribution 
$\hat D(u)$. This is the bootstrap computation, which is visualized in Figure~\ref{fig:actual}.

The theory of bootstrap 
guarantees that for a large class of estimators $\theta$ and sufficiently large $k$, we can
use $\hat D(u)$ as a consistent approximation of $D(u)$. 
The intuition is that, by resampling from $L$, we emulate the original distribution $\mathbb{D}$ that generated $L$.
Here, it is sufficient (but not necessary) that $\theta$ be relatively smooth 
(i.e., Hadamard differentiable~\cite{efron93bootstrap}) which holds for a large class of machine learning
 algorithms~\cite{M_estim_bootstrap} such as $M$-estimators, themselves
including maximum   likelihood estimators and most classification techniques.
In our experiments (Section~\ref{sec:expr}), $k$=$100$ or even $10$ have yielded reasonable accuracy ($k$ can also be tuned automatically; see~\cite{efron93bootstrap}).


Both of our AL algorithms use bootstrap to estimate the classifier's uncertainty in its predictions (say, to stop asking the crowd 
once we are confident enough).
Employing bootstrap has several
key advantages. First, as noted, bootstrap
delivers consistent estimates for a large class
 of estimators, making our AL algorithms general and applicable to nearly any classification algorithm.\footnote{
The only known exceptions are lasso classifiers (i.e., L1 regularization).
 Interestingly, even for lasso classifiers, there is a modified version of bootstrap that can produce consistent estimates \cite{bootstrap-lasso}.}
Second, the bootstrap computation uses  a ``plug-in'' principle; that is,  
we simply need to invoke our estimator $\theta$ with $S_i$ instead of $L$. 
Thus, we can  treat $\theta$ (here, our classifier)
as a complete black-box since its internal implementation does not need to be modified.
Finally, individual bootstrap computations $\theta^{S_1}(u),\cdots,\theta^{S_k}(u)$ 
 are independent from each other, and 
 hence can be executed in
parallel. This embarrassingly parallel execution model  enables scalability by taking full advantage 
  of modern many-core and distributed 
systems.

Thus,
by exploiting powerful theoretical results from classical nonparametric statistics,
we can estimate the uncertainty of complex estimators and also scale up the computation to large volumes of data. 
Aside from Provost et al.~\cite{provost} (which is limited to probabilistic classifiers and is less effective than our algorithms; see Sections \ref{sec:expr} and \ref{sec:related}), no one has exploited the power of bootstrap  in AL, perhaps due to bootstrap's computational overhead. However, with recent advances in parallelizing and optimizing bootstrap computation~\cite{mozafari_sigmod2014_diagnosis,blb,mozafari_sigmod2014_demo,mozafari_sigmod2014_abm} and  increases in RAM sizes and the number of CPU cores, bootstrap is now a computationally viable approach, motivating our use of it in this paper.

\subsection{\un Algorithm}
\label{sec:uncertainty}

Our \un algorithm aims to ask the crowd the questions that are hardest for the classifier. 
Specifically, we (i)  
find out how uncertain (or certain) our given classifier $\theta$ is 
in its label predictions for 
 different unlabeled items, and (ii) ask the crowd to label  items for which 
 the classifier is least certain. 
  The intuition is that, the more uncertain the  classifier, 
 the more likely it will mislabel the item.
 
Note that focusing on most uncertain items is 
one of the oldest ideas in AL literature~\cite{jordan-96, svm-margin-old}. 
The novelty and power of our \un algorithm is in its use of bootstrap for obtaining an \emph{unbiased} 
estimate of uncertainty, while making almost no assumptions about the classification algorithm.
Previous proposals that capture uncertainty 
 either (i) require a {\it probabilistic} classifier that can produce highly accurate
class probability estimates along with its label predictions \cite{provost, vision-al}, or
 (ii) are limited to a particular classifier.
For example, 
the 
standard way of performing uncertainty-based sampling is by using the entropy of the class distribution
 in the case of probabilistic classifiers~\cite{raykar2011entropic, zhou2012learning}. 
Although entropy-based AL can be effective in some situations \cite{fan2014hybrid,raykar2011entropic, zhou2012learning},  
in many other situations,  
when the classifiers do not produce accurate probabilities,  
   entropy does not guarantee an unbiased 
  estimate of the uncertainty (see Section \ref{sec:uci}).
  For non-probabilistic classifiers, other heuristics such as the distance from the separator is taken as a measure of uncertainty (e.g., SVM classifiers \cite{svm-margin, svm-entropy}). 
 However,  these heuristics cannot be applied to arbitrary classifiers.
 In contrast, our \un algorithm  applies to both  probabilistic and non-probabilistic classifiers, and is also guaranteed by
 bootstrap theory to produce unbiased estimates. 
(In Section~\ref{sec:expr}, we also
empirically show that  our  algorithm is more effective.) 
Next, we describe how \un uses bootstrap to estimate the classifier's uncertainty.


Let $l$ be the predicted  label for item $u$ when we train $\theta$ on our labeled data $L$, i.e., $\theta^L(u)=l$.
As explained in Section~\ref{sec:bootstrap}, $L$ is often a random variable, and hence, $\theta^L(u)$ has a distribution (and variance). 
We use the variance of $\theta$  in its prediction, namely $Var[\theta^L(u)]$, as our formal notion of \emph{uncertainty}. 
Our intuition behind this choice is as follows. 
A well-established result from Kohavi and Wolpert~\cite{Kohavi96biasplus} has shown that the classification error for item $u$, say $e_u$,
 can be decomposed into a sum of three terms:
 $$
 e_u = \var[\theta^L(u)] + bias^2[\theta^L(u)] + \sigma^2(u)
 $$
where $bias[.]$ is  the  bias of  the classifier
and $\sigma^2(u)$ is  a noise term.\footnote{Squared bias $bias^2[\theta^L(u)]$ is defined as 
$[f(u)-E[\theta^L(u)]]^2$, where $f(u)$=$E[l_u|u]$, i.e.,
 expected value of  true label given $u$ \cite{Kohavi96biasplus}.}
 Our ultimate goal in AL is to reduce the sum of $e_u$ for all  $u$.
 The
$\sigma^2(u)$  is an error inherent to the data collection process,
and thus cannot be eliminated through AL. 
 Thus, 
 by requesting labels for $u$'s that have a large variance, we indirectly reduce  
  $e_u$ for $u$'s that have a large classification error.\footnote{Note that we could try to choose items based on the bias
  of the classifier. 
  In Section~\ref{sec:ete}, we present an algorithm that indirectly reduces both variance and bias in a mathematically sound way.}
Hence, our \un algorithm assigns  $\var[\theta^L(u)]$ as the score for each unlabeled item $u$, to
ensure that  items with larger variance are sent to the crowd for labels.
 Thus, in \un algorithm, our goal is to measure  $\var[\theta^L(u)]$.

\begin{figure*}[t]
\vspace{-0.2in}
\begin{minipage}{3.2in}
\centering
\hspace{-0.1in}
\subfigure[Ideal computation of $D(u)$]{\label{fig:ideal} \includegraphics[height=0.8in]{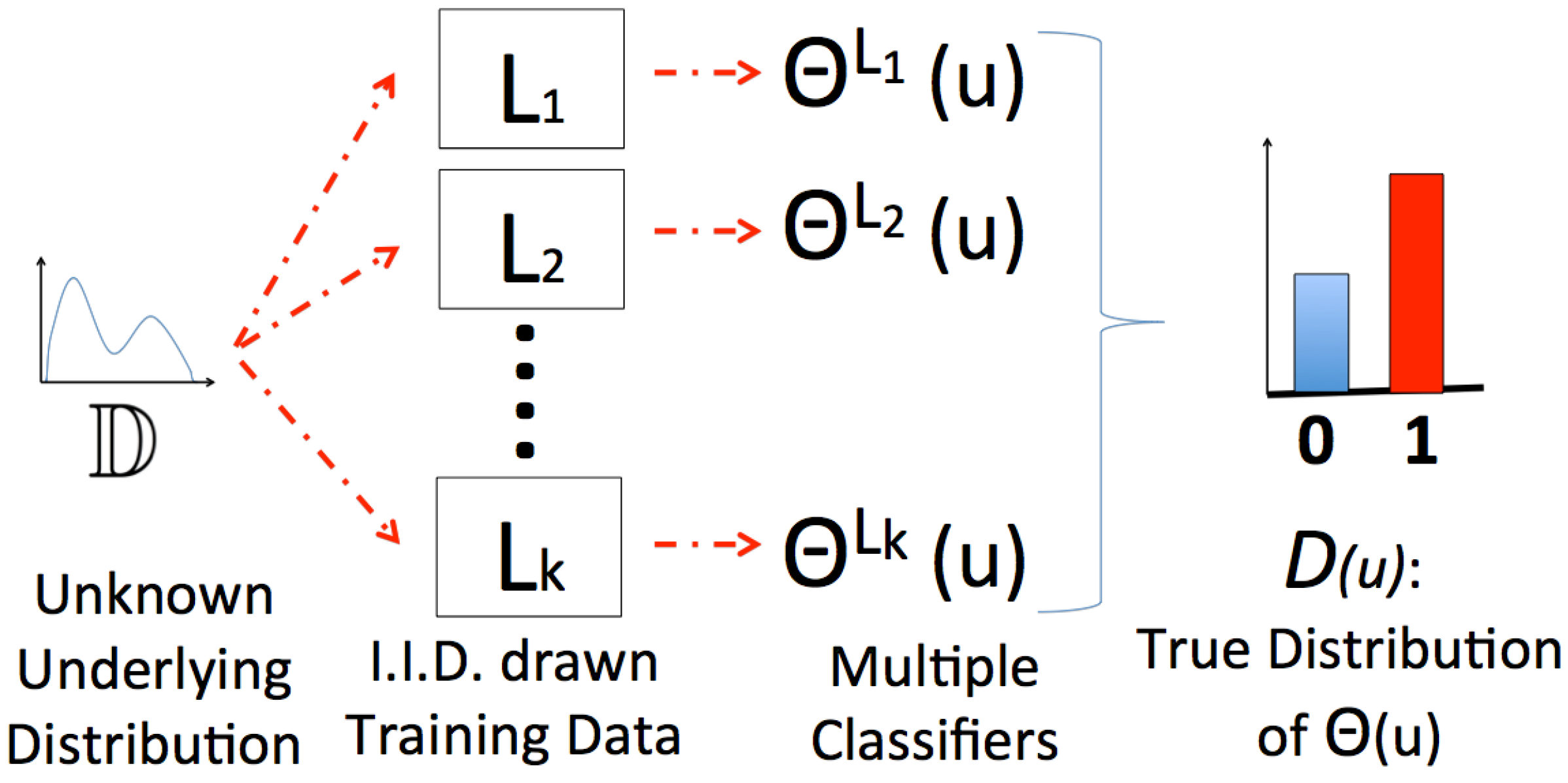}}
\hspace{0.05in}
\subfigure[Bootstrap computation]{\label{fig:actual} \includegraphics[height=0.76in]{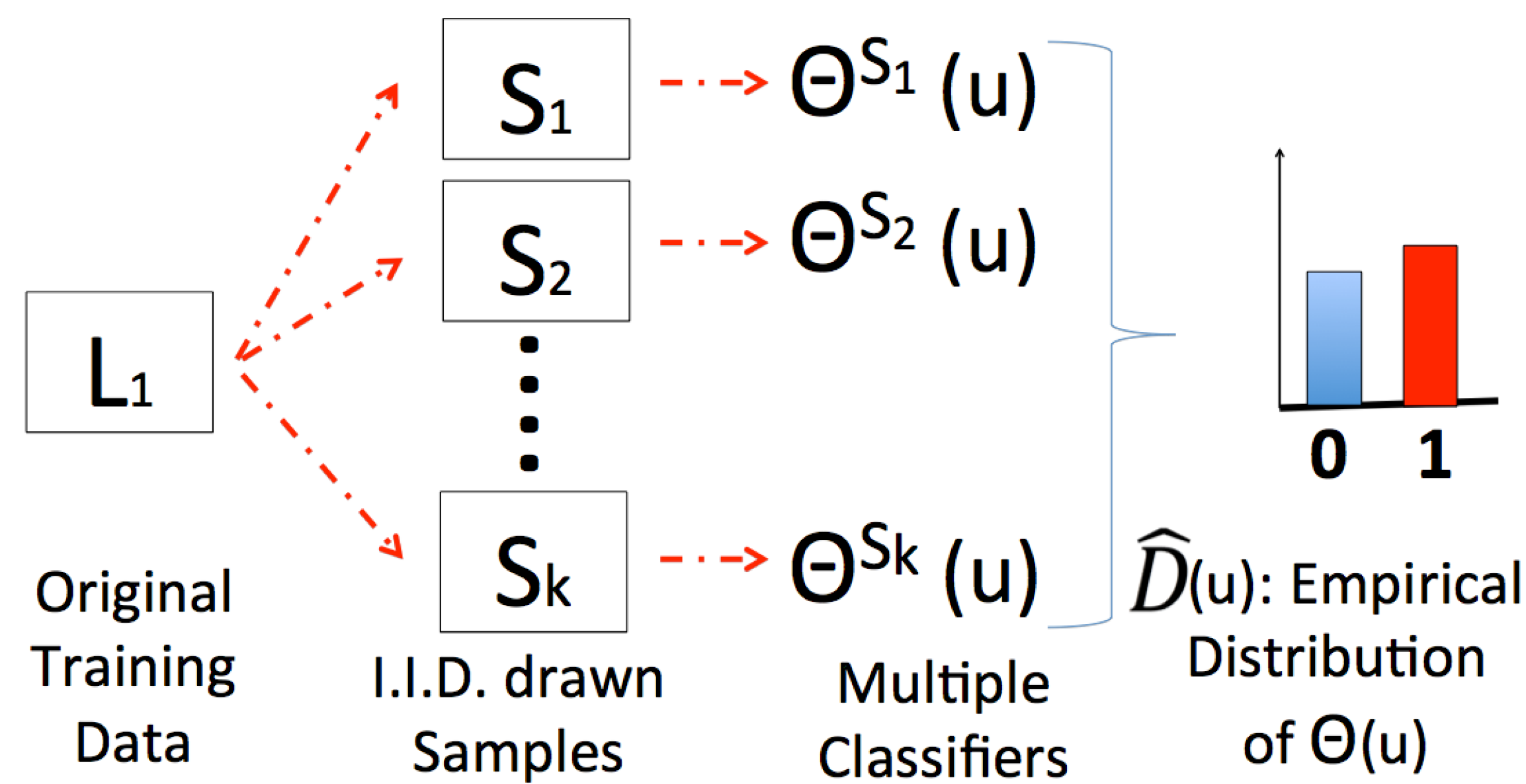}}
\caption{{\small Bootstrap approximation $\hat D(u)$ of true distribution $D(u)$.}}
\label{fig:bootstrap}
\end{minipage}
\hspace{0.3in}
\begin{minipage}{3.6in}
\centering
\hspace{-0.4in}
\subfigure{\label{fig:ex:full} \includegraphics[height=1.3in]{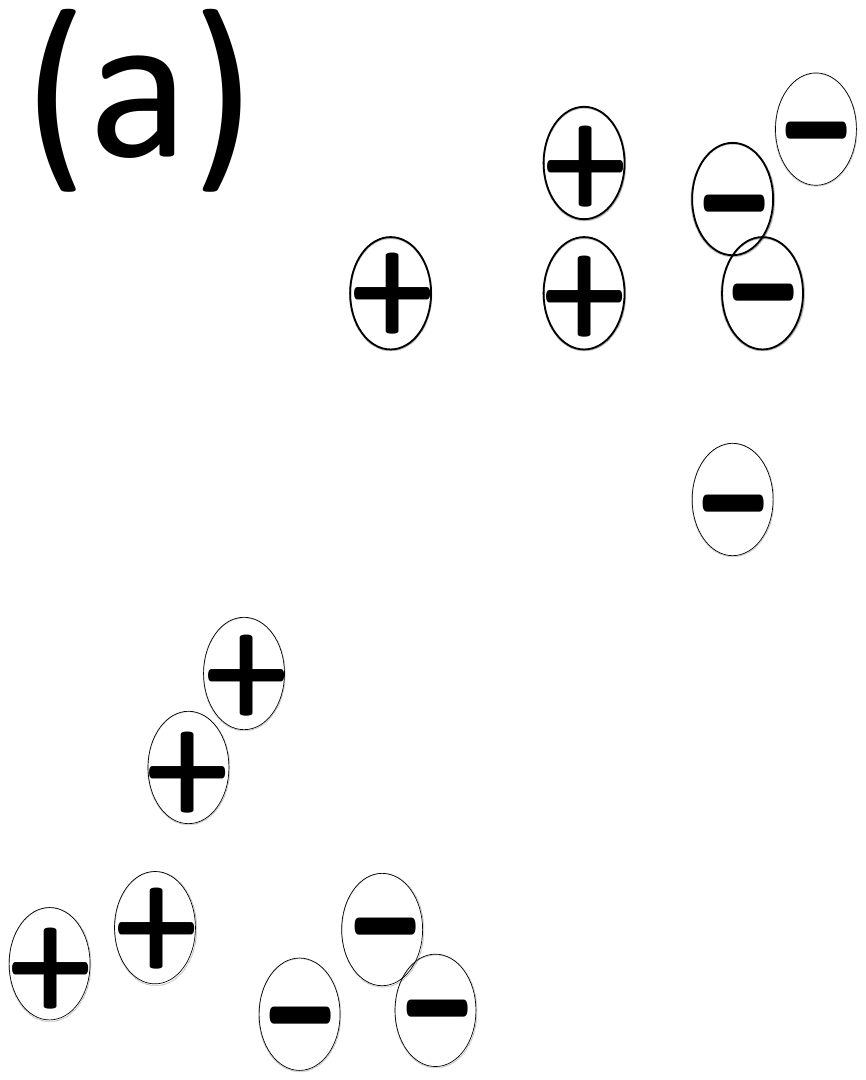}}
\hspace{-0.3in}
\subfigure{\label{fig:ex:initial} \includegraphics[height=1.3in]{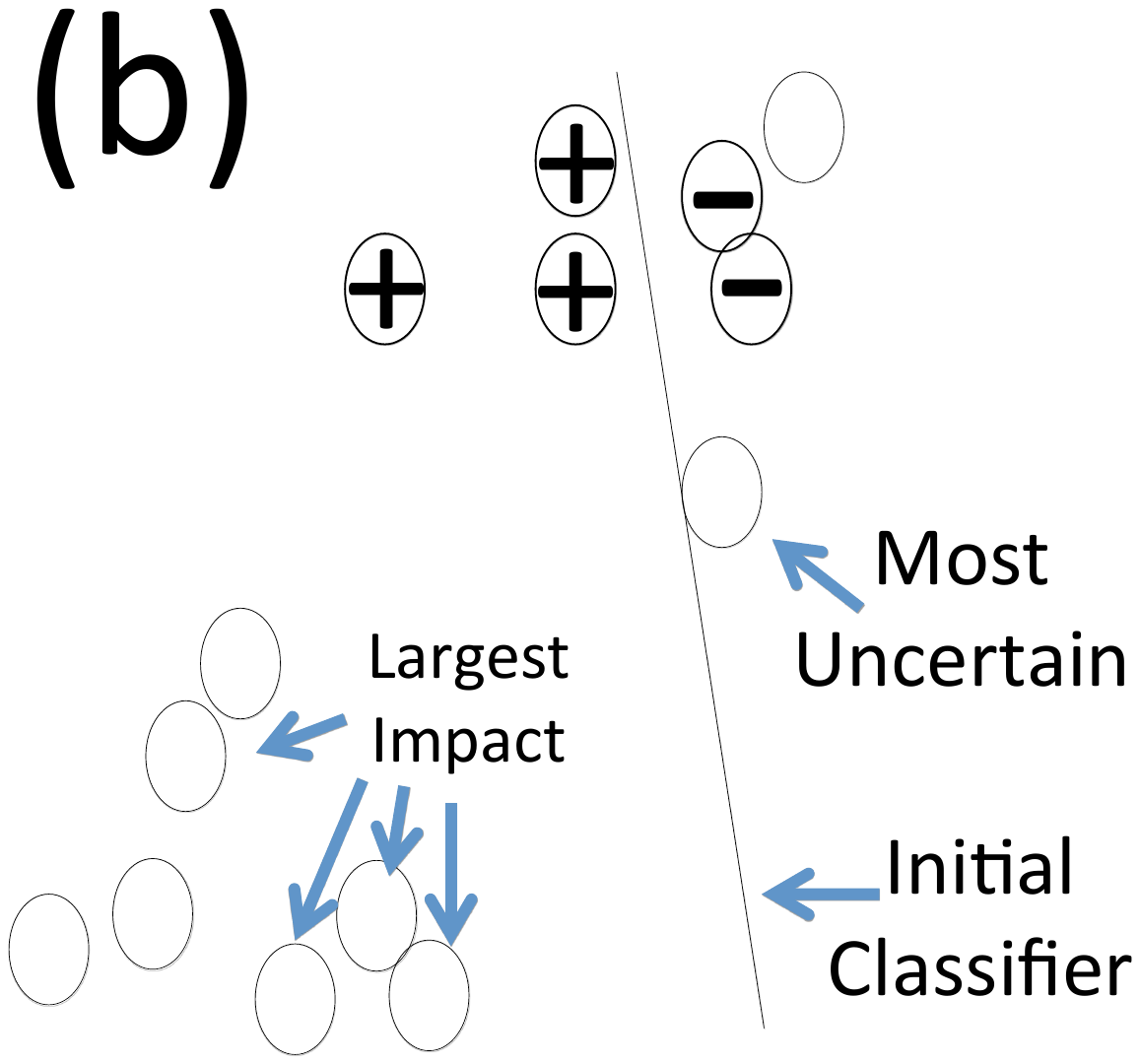}}
\hspace{-0.1in}
\subfigure{\label{fig:ex:uncertainty} \includegraphics[height=1.3in]{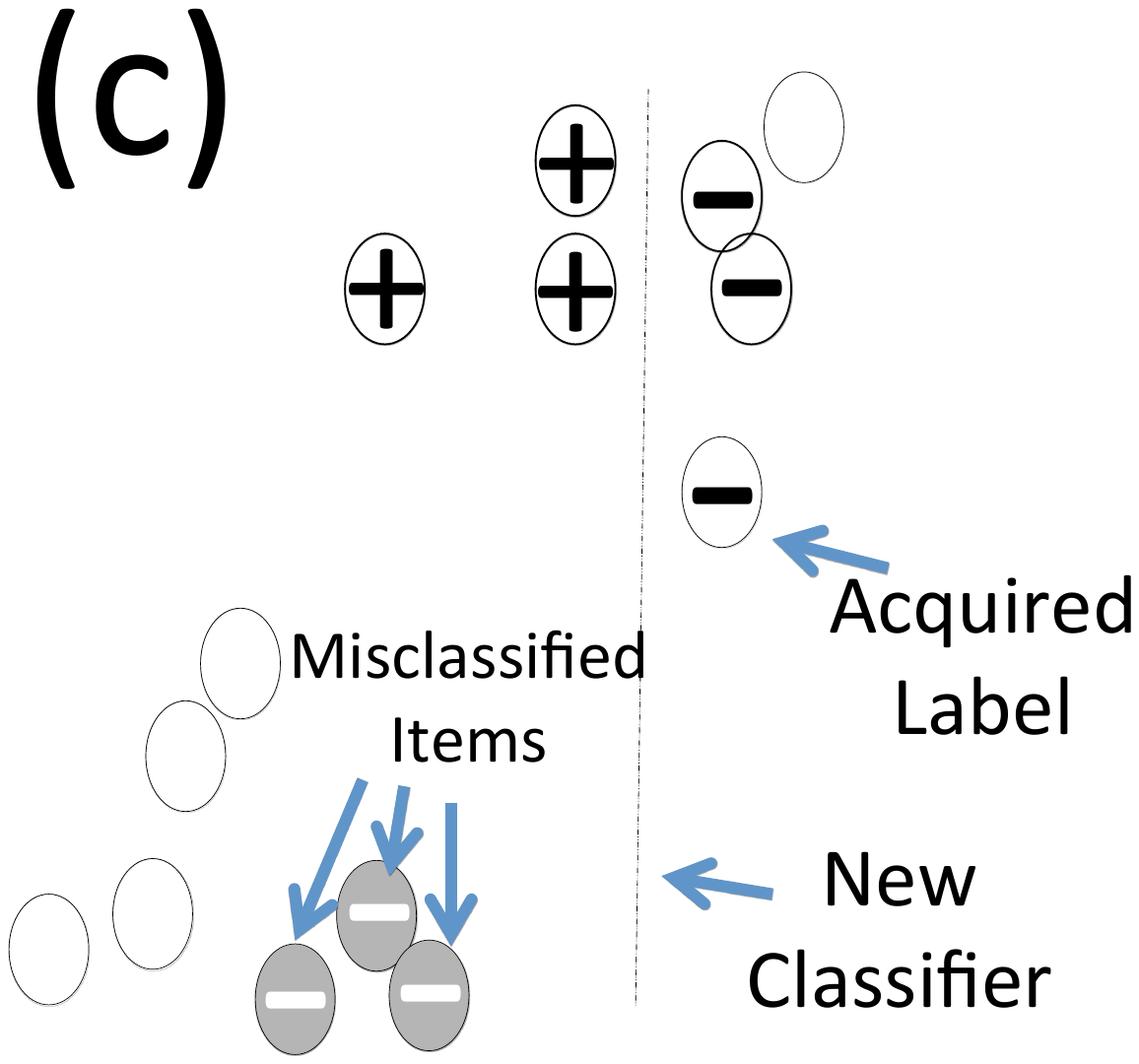}}
\hspace{-0.1in}
\subfigure{\label{fig:ex:impact} \includegraphics[height=1.3in]{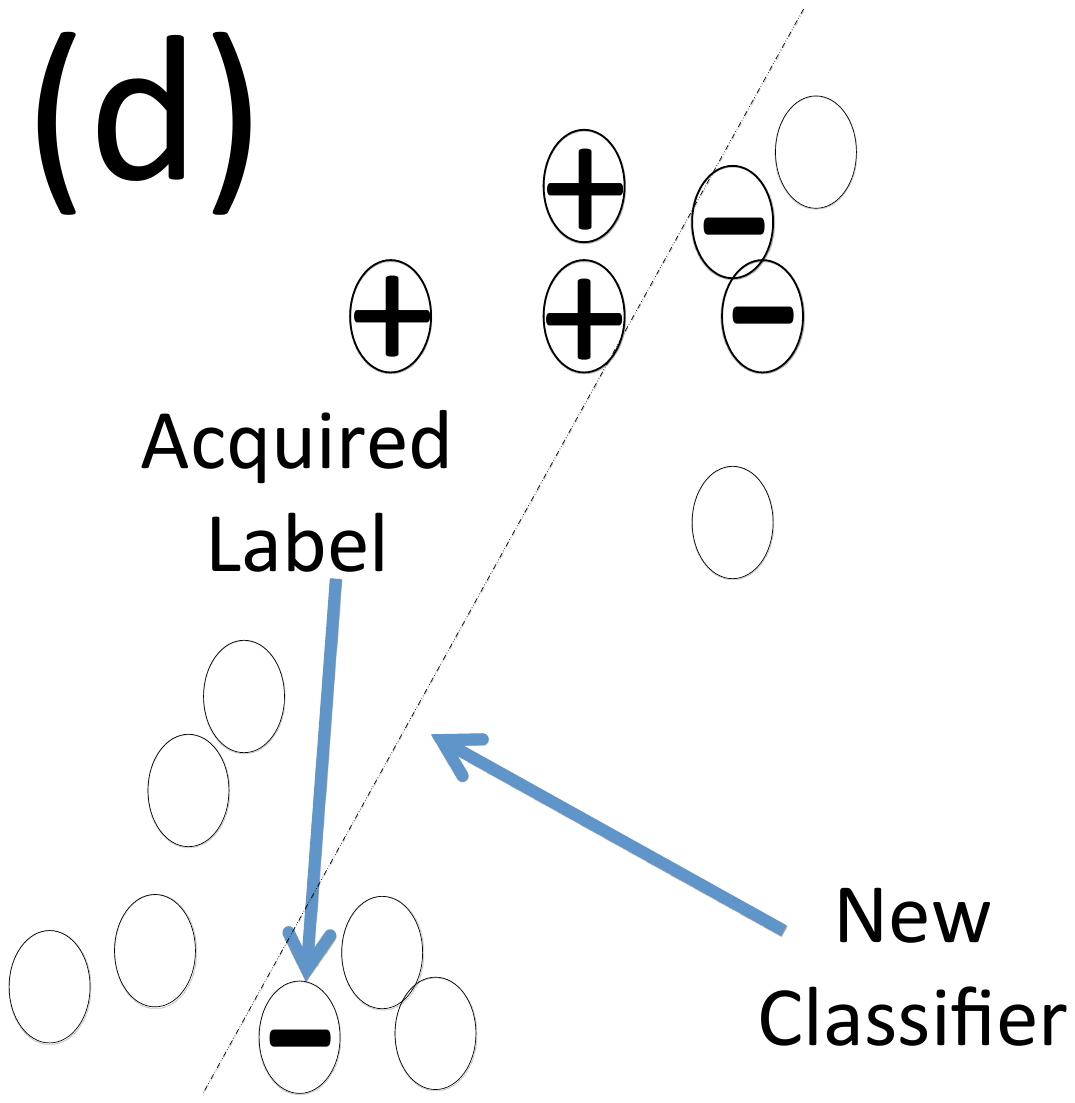}}
\hspace{-0.3in}
\vspace{-0.1in}
\caption{{\small (a) Fully labeled dataset, (b) initial labels, (c) asking hardest \newline questions, and (d) asking questions with high impact.}}
\label{fig:ex:algorithm}
\end{minipage}
\end{figure*}

 Since the underlying distribution of the training data ($\mathbb{D}$ in Figure~\ref{fig:bootstrap}) is unknown to us,
 we use bootstrap. 
 In other words,
we bootstrap our current set of labeled data $L$, say  $k$ times,  to obtain $k$ different classifiers
that are then invoked to generate labels for each item $u$. This is shown in Figure~\ref{fig:actual}. 
The output of these classifiers form
 an  \emph{empirical distribution} $\hat D(u)$ that approximates the true distribution of  $\theta^L(u)$.
   We can then estimate  $\var[\theta^L(u)]$ using $\hat D(u)$ which is guaranteed, by bootstrap theory~\cite{efron93bootstrap},  to quickly converge to the true value of $\var[\theta^L(u)]$ as we increase $k$.
          
Let $S_i$ denote the $i$'th bootstrap, and $\theta^{S_i}(u)=l^i_u$ be the prediction of our classifier for $u$
when trained on this bootstrap.
Define $X(u):=\sum_{i=1}^{k} l^i_u/k$, i.e., the fraction of classifiers in Figure~\ref{fig:actual} that predict a label of $1$ for $u$.
Since $l^i_u\in \{0,1\}$, the uncertainty score for instance $u$ is given by its variance, which can be computed as:
\begin{equation}
    \un(u) = \var[\theta^L(u)] = X(u)(1-X(u))
\end{equation}
We evaluate our \un algorithm in Section~\ref{sec:expr}.

\subsection{\ete Algorithm}
\label{sec:ete}

Consider the toy dataset of Figure~\ref{fig:ex:full}. Initially, only a few labels ($+$ or $-$) are revealed to us,  as
shown in Figure~\ref{fig:ex:initial}. With these initial labels we train a classifier, say a linear separator (shown as a solid line).
The \un algorithm would ask the crowd to label the items with the most uncertainty, here being those closer to the 
separator.  The intuition behind  \un  is that, by measuring the \textit{uncertainty} 
 (and requesting labels for the most uncertain items),
the crowd essentially handles items that are \textit{hardest} (or ambiguous) for the classifier.
However, this might not always be  the best strategy. 
By acquiring a label for the item closest to the separator and training a new classifier, as shown in Figure~\ref{fig:ex:uncertainty}, our overall accuracy does not change much: despite acquiring a new label, the new classifier still misclassifies 
three of the items at the lower-left corner of Figure~\ref{fig:ex:uncertainty}. This observation shows that 
labeling items with most uncertainty (i.e., asking the hardest questions) may not  have the \emph{largest impact} on the classifier's prediction power 
for other data points. 
In other words, another 
great strategy would be to acquire human labels for items that,
if their labels differ from what the current classifier thinks, 
would have a huge impact on the classifier's future decisions.
In Figure~\ref{fig:ex:impact},  the lower-left points exemplify such items. 
We say that such items have a  \emph{potential for largest impact} on the classifier's accuracy.

Note that we cannot completely ignore  uncertainty and  choose items only based on their \emph{potential} impact on the classifier. When the classifier is highly confident of its predicted label, no matter how much impact 
an opposite label could have on the classifier, acquiring a label for that item wastes resources  
because the crowd label will most likely agree with that of  the classifier anyway.    
Thus, our \ete algorithm  combines these two strategies in a mathematically sound way, 
described next.



Let $l=\theta^L(u)$ be the current classifier's predicted label for $u$.
If we magically knew that $l$ was the correct label, 
 we could simply add $\langle u,l\rangle$ to $L$ and retrain the classifier. 
 Let $e_{\textrm{right}}$ be this new classifier's error. 
 On the other hand, if we somehow knew that $l$ was the incorrect label,
 we would instead add  $\langle u,1-l\rangle$ to $L$ and retrain the classifier accordingly. 
 Let $e_{\textrm{wrong}}$ denote the error of this new classifier. 
  The problem is that (i) we do not know what the true label is, and (ii) we do not know the classifier's error
   in  either case.
   
 Solving (ii) is relatively easy: in each case we assume those labels and use cross validation on $L$ 
 to estimate both errors, say $\hat e_{\textrm{right}}$ and
 $\hat e_{\textrm{wrong}}$.
  To solve  problem (i), we can again use bootstrap to estimate the probability of our prediction $l$ being correct (or incorrect),
 say $p(u):=Pr[l=l_u|u]$, where $l_u$ is $u$'s  true label. 
Since we do not know $l_u$ (or its distribution),
we bootstrap $L$ to train
 $k$ different classifiers (following Figure~\ref{fig:bootstrap}'s notation). 
 Let $l_1,\cdots,l_k$ be the labels predicted by these classifiers for $u$. 
 Then, $p(u)$ can be approximated as 
\begin{equation}
     \hat{p}(u) = \frac{\sum_{i=1}^{k} 1(l_{i} = l)}{k}
\label{eq:ete}
\end{equation}
Here, $1(c)$ is the decision function which evaluates to $1$ when condition $c$ holds and to zero otherwise.
Intuitively, equation (\ref{eq:ete}) says that the probability of the classifier's prediction being correct can 
be estimated by the fraction of  
classifiers that agree on that prediction, {\it if} those classifiers are each trained on a bootstrap of the training set $L$.


Our \ete algorithm aims to  
compute the classifier's expected error if we use 
the classifier itself to label each item, and ask the crowd to label those items
for which this expected error is larger. This way, the overall expected error will be minimized.
To compute the classifier's expected error, we can average over different label choices:
\begin{eqnarray}
\ete(u) & = & \hat{p}(u) \hat e_{\textrm{right}}+(1-\hat{p}(u))\hat e_{\textrm{wrong}}
\label{eq:mexperr}
\end{eqnarray}

 We can break down equation (\ref{eq:mexperr}) as:
\begin{eqnarray}
\ete(u)& = & \hat e_{\textrm{wrong}}-\hat{p}(u)(\hat e_{\textrm{wrong}}-\hat e_{\textrm{right}})\label{eq:ete2}
\end{eqnarray}

Assume that $\hat e_{\textrm{wrong}}-\hat e_{\textrm{right}}\geq 0$ (an analogous decomposition is possible when it is negative).
Eq. (\ref{eq:ete2}) tells us that if the question is too hard (small $\hat{p}(u)$), 
we may still  ask for a crowd label to avoid a high risk of misclassification on $u$.
On the other hand, we may ask a question for which our model is fairly confident (large $\hat{p}(u)$), but having its true label can still make a big difference in classifying other items ($\hat e_{\textrm{wrong}}$ is too large). 
This means that, however unlikely, if our classifier happens to be wrong, we will have a higher overall error if we do \emph{not} ask for the true label of $u$. Thus, our \ete scores naturally combine
both the difficulty of the question and how much knowing its answer can improve our classifier.




\subsection{Complexity and Scalability}
\label{sec:scalability}

Besides its generality, a major benefit of bootstrap is that
each  replicate can be shipped to a different node or processor, performing training in parallel. 
The time complexity of each iteration of  \un  is $O(k\cdot T(|U|))$, where $|U|$ is the number of unlabeled items
 in that iteration, $T(.)$ is the classifier's training time (e.g., this is cubic in the input size for SVMs),
and $k$ is the number of bootstraps. Thus, we only need $k$ nodes to achieve the same run-time as training a single classifier. 

\ete is  more expensive than \un as it requires a case analysis for each unlabeled item.
The time complexity for each iteration of  \ete  is $O((k+|U|)\cdot T(|U|))$. 
Since unlabeled items can be independently analyzed, the algorithm is still parallelizable.
However, \ete  requires more nodes (i.e., $O(|U|)$ nodes) to achieve the same performance as \un. 
As we show in Section~\ref{sec:expr}, 
this additional overhead is justified in the upfront scenario, given \ete's superior performance at  
deciding which questions to ask based on a limited set of initially labeled data.


\if{0}
\subsection{Theoretical Concerns}
\label{sec:limitations}

\question{Talk about violations of IID-ness}

\barzan{As mentioned in Section~\ref{sec:intro}, it is common for AL algorithms to make assumptions 
that greatly simplify theoretical analysis. 
For instance, a streaming setting (where unlabeled items are revealed one at a time) allows for use of IID-based results~\cite{iwal, iwalbb, general-agnostic}. 
Similarly. having access to the current hypothesis space of the classifier and being able to shrink it by adding new constraints~\cite{iwal}, or
relying on VC-dimension~\cite{ml-textbook} bounds allow for provable \emph{statistical consistency} and \emph{label complexity}\footnote{Informally, statistical consistency guarantees that 
the AL will eventually reach the best possible model, and label complexity bounds the number of labels required.}\cite{para-active, iwalbb}.
However, as explained in Section~\ref{sec:intro}, these assumptions do not lend themselves to a practical and scalable crowd-sourced query processor 
that needs to be used by many practitioners from different domains.}

\question{should we move parts of the following to the intro?}
\barzan{Hence, in our pursuit of achieving generality, scalability (batching and parallelism), and ease-of-use (black-box view and automatic noise-management) 
as our guiding principles,
we have traded the simplifying assumptions often made for 
theoretical analysis, in exchange for \emph{sound heuristics} that 
have enabled us to build a practical system which we have empirically  
evaluated across a wide range of applications (see Section~\ref{sec:experiments}).
We do not claim that provable guarantees cannot be achieved along with these designs goals. 
On the contrary, we believe that it is critical to re-examine the theoretical AL frameworks in
light of new system challenges posed by the recent rise of crowd-sourcing platforms and publicly available datasets that are growing in size.
We hope that the solution proposed in this paper will serve as a starting point for the adoption of AL strategies 
in data-intensive systems, and at the same time, encouraging system-oriented criteria in the design of the next generation of AL algorithms.} 

\fi


\section{Handling Crowd Uncertainty}
\label{sec:noise}

Crowd-provided labels are subject to a great degree of uncertainty: 
humans may give incorrect answers due to ambiguity of the question and innocent (or deliberate) errors.
 This section proposes an algorithm called {\it Partitioning Based Allocation (\pba)} that manages
and reduces this uncertainty
 by strategically  allocating different degrees 
 of redundancy to different subgroups of the unlabeled items. 
\pba is our proposed instantiation of $\Gamma$ in the upfront and iterative scenarios (Figures \ref{fig:upfront} and \ref{fig:iterative})
which given a fixed budget $B$ maximizes the labels' accuracy, or given a required accuracy, minimizes cost.

\vspace*{0.1in}
\noindent{\bf Optimizing Redundancy for Subgroups.} Most previous AL approaches 
assume that labels are provided by domain experts
and thus perfectly correct (see Section~\ref{sec:related}).  
In contrast,  incorrect labels are common in a crowd database ---
an issue  conventionally handled by using
{\it redundancy}, e.g.,
asking each question to multiple workers and combining their answers for the best overall
result.
Standard techniques, such as asking for multiple answers
and using majority vote or the techniques of Dawid and Skene
(DS)~\cite{dawid} can improve answer quality when the crowd is mostly
correct, but 
will not help much if users do not converge to the right answer or converge too slowly.  In our experience, 
crowd workers can be quite imprecise for certain classification tasks.
For example, we removed the labels from $1000$ tweets with hand-labeled (``gold 
data'') sentiment (dataset details in Section~\ref{sec:twitter}), and asked Amazon Mechanical Turk workers to label them
again, then measured the workers' agreement.  
We used different redundancy
ratios ($1, 3, 5$) and different voting schemes (majority and DS) and computed the  crowd's ability to agree with
the hand-produced labels. The results are shown in
Table~\ref{tab:agreement}.

\begin{table}[th]
\vspace{0.1in}
\centering
 \begin{tabular}{|m{2.7cm}|m{1.9cm}|m{1.9cm}|m{2.5cm}|}
 \hline
Voting Scheme  & Majority Vote &  Dawid~\&~Skene\\
 \hline
 \hline
 1 worker/label & 67\verb=%= & 51\verb=%= \\
 \hline
 3 workers/label & 70\verb=%= & 69\verb=%= \\
 \hline
 5 workers/label & 70\verb=%= &  70\verb=%= \\
 \hline
\end{tabular}
\caption{{\small The effect of redundancy (using both majority voting and Dawid and Skene voting) on the accuracy of crowd labels.}}
\vspace{0.1in}
\label{tab:agreement}
\end{table}

In this case,
increasing redundancy from $3$ to $5$ labels
  does not significantly increase the
crowd's accuracy.
Secondly, we have noticed that crowd accuracy varies for different subgroups of the unlabeled data.  For example,
in a different experiment,
we asked Mechanical Turk workers to label facial expressions in the CMU Facial Expression 
dataset,\footnote{http://kdd.ics.uci.edu/databases/faces/faces.data.html}
and measured  agreement with hand-supplied labels. This dataset consists of $585$ head-shots of 
$20$ users, each in $32$ different combinations of head positions (straight, left, right, and up), sunglasses (with and without), and facial expressions (neutral, happy, sad, and angry).
The crowd's accuracy was significantly worse
when the faces were looking up versus  other positions:


\vspace{0.2cm}
\begin{center}
 \begin{tabular}{|m{2.4cm}|m{1.9cm}|}
 \hline
 \textbf{Facial orientation} & \textbf{Avg. accuracy} \\
 \hline
 \hline
  straight & 0.6335\verb=%= \\
 \hline
  left & 0.6216\verb=%= \\
 \hline
  right & 0.6049\verb=%= \\
 \hline
  up & 0.4805\verb=%= \\
 \hline
 \end{tabular}
 \end{center}
\vspace{0.15cm}

Similar patterns appear in several other datasets, where crowd accuracy is considerably lower for certain subgroups.
To exploit these two observations, we developed our \pba algorithm which 
computes the optimal number of questions to ask about each subgroup by estimating 
the probability $p_g$ with which the crowd correctly classifies items of a given subgroup $g$, and then solves an integer linear program (ILP) to choose the optimal number of questions (i.e., degree of redundancy) for labeling each item from that subgroup, given these probabilities.

Before introducing our algorithm, we make the following observation. 
To combine answers using majority voting and an odd number of votes, say $2v + 1$, for an unlabeled item $u$ with a true label $l$,
the probability of the crowd's combined answer $l^*$ being correct
is the probability that at most $v$ or fewer workers get the answer wrong.
Denoting this probability with $P_{g,(2v+1)}$, we have:
\begin{eqnarray}
\vspace*{-0.1in}
P_{g,(2v+1)} = Pr(l = l^* | 2v+1\text{ votes}) = \nonumber \\
\sum_{i=0}^v\binom{2v+1}{i}\cdot p_g^{2v+1-i} \cdot (1-p_g)^i
\label{eq:binomial}
\end{eqnarray}
where $p_g$ is the probability that a crowd worker will correctly label an item in group $g$.

Next, we describe our \pba 
algorithm, which partitions the items into subgroups
and optimally allocates the budget to different subgroups by computing the optimal 
number of
votes per item, $V_g$, for each subgroup  $g$.
 \pba consists of three steps:


\vspace{0.1in}
\textbf{Step 1.} Partition the dataset into $G$ subgroups.  This can be done either by 
partitioning on some low-cardinality field that is already present
 in the dataset to be labeled (for example, in an image recognition 
dataset, we might partition by photographer ID  or the 
time of day when the picture was shot), or by using an 
unsupervised clustering algorithm such as $k$-means.
For instance, in the CMU facial expression dataset, we partitioned the images based on user IDs, leading to $G=20$ subgroups, each with roughly $32$ images.

\vspace{0.1in}
\textbf{Step 2.}  Randomly pick $n_{0}$$>$$1$ different data items from each subgroup, and obtain $v_{0}$ labels for each one of them.
 Estimate $p_g$ for each subgroup $g$, either by choosing data items for which the label is known and computing the fraction of labels that are correct, or by taking the majority vote for each of the $n_{0}$ items, assuming it is correct, and then computing the fraction of labels that agree with the majority vote.
 For example,
for the CMU dataset, we asked for $v_{0}=9$ labels for $n_{0}=2$ random images\footnote{With a similar cost, 
we could ask for $v_{0}$$=$$5$ labels for $n_{0}$$=$$4$ images per group.
Using  larger $v_0$ and $n_0$ (and accounting for each worker's accuracy) will yield more reliable  estimates for  $p_g$'s. 
For CMU dataset, we use a small budget to show that even with  rough estimates we can 
easily improve on uniform allocations. We study the effect of $n_0$ on \pba's performance in Section \ref{s:pba-expts}.} from each subgroup, and hand-labeled those $n_{0}$$*$$G=40$ images to estimate $p_g$ for $g=1,\dots,20$.

\vspace{0.1in}
\textbf{Step 3.}  Solve an ILP  to find the optimal $V_g$ for every group $g$.
We use $b_g$ to denote the budget allocated to subgroup $g$, and create a binary indicator variable $x_{gb}$ whose value is 1 iff subgroup $g$ is allocated a budget of $b$.  Also,
let $f_g$ be the number of items that our leaner has chosen to label from subgroup $g$. 
Our ILP formulation depends on the user's goal: 

\vspace{0.1in}
\textbf{Goal 1.} Suppose we are given a budget $B$ (in terms of the number of questions) and our goal is to 
acquire the most accurate labels for the items requested by the learner.  
We can then formulate an ILP to minimize the following objective function:
\begin{equation}
 \sum_{g=1}^{G} \sum_{b=1}^{b^{max}} x_{gb} \cdot (1-P_{g,b}) \cdot f_g
 \label{eq:goal}
\end{equation}
where $b^{max}$ is the maximum number of votes that we are willing to ask per item.
This goal function captures the expected weighted error of the crowd, 
 i.e., it has a lower
value when we allocate a larger budget ($x_{gb}$=$1$ for a large $b$ when $P_{g}$$>$$0.5$) to subgroups whose questions are harder for the crowd
 ($P_{g,b}$ is small) or the learner has chosen more items from that group ($f_g$ is large).
This optimization is subject to the following constraints:
\begin{eqnarray}
\forall 1\leq g \leq  G, ~~~ \sum_{b=1}^{b^{max}}  x_{gb} = 1  \label{eq:c1} \\
 \sum_{g=1}^{G} \sum_{b=1}^{b^{max}} x_{gb} \cdot b \cdot f_g \leq B - v_0 \cdot n_0 \cdot G\label{eq:c2}
\end{eqnarray}

Here,  constraint (\ref{eq:c1}) ensures that we pick exactly one $b$ value for each subgroup and  (\ref{eq:c2}) ensures that we  stay within our \ labeling budget (we subtract the initial cost of estimating $p_g$'s from $B$).

\vspace{0.1in}
\textbf{Goal 2.} If we are given a minimum required accuracy  $Q$, and our goal is to minimize the total number of questions asked, 
we modify the formulation above by turning (\ref{eq:c2}) into a goal and (\ref{eq:goal}) into a constraint, i.e., minimizing 
$ \sum_{g=1}^{G} \sum_{b=1}^{b^{max}} x_{gb} \cdot b \cdot f_g$ while ensuring that 
$ \sum_{g=1}^{G} \sum_{b=1}^{b^{max}} x_{gb} \cdot (1-P_{g,b}) \cdot f_g \leq 1- Q$.

Note that one could further improve the $p_g$ estimates and the expected error estimate in (\ref{eq:goal}) by incorporating the
 past accuracy of individual workers \cite{bs2}. Such extensions are omitted here due to lack of space and left to future work.
We evaluate  \pba in Section~\ref{s:pba-expts}.

\vspace{.1in}
\noindent{\bf Balancing Classes.} Our final observation about the crowd's accuracy is that crowd workers perform better at classification tasks 
when the number of instances from each class is relatively balanced.  For 
example, given a face labeling task asking the crowd to tag each face
 as ``man'' or ``woman'' where only $0.1\verb=%=$ of images are of men,
 crowd workers will have a higher error rate when labeling men (i.e.,  the rarer class). 
Perhaps workers become conditioned to answering ``woman''.  
(Psychological studies report the same effect~\cite{wolfe-rare}.)

Interestingly, both our \un and \ete algorithms naturally tend to
increase the fraction of labels they obtain for rare classes.  
Since  our algorithms  tend to have more uncertainty about items with rare
labels (due to insufficient examples in their training set), 
they are more likely to ask users to label those items.  Thus, our
algorithms naturally improve the ``balance'' in the questions they ask
about different classes, which in turn improves crowd labeling
performance.  We show this effect in Section~\ref{s:pba-expts} as well.

\section{Optimizing for the Crowd}
\label{sec:optimization}

The previous section described our algorithm for handling noisy  labels.
There are other optimization
questions that arise in practice.
How should we  decide when our accuracy is ``good enough''? (Section~\ref{sec:stop})
Given that a crowd can label multiple items in parallel, 
what is the   the effect of batch size (number of simultaneous questions) on our learning performance? (Section~\ref{sec:batch})  
 
 \subsection{When To Stop Asking}
\label{sec:stop}

As mentioned in Section~\ref{sec:regimes}, users may either provide a fixed budget $B$ or a minimum
quality requirement $Q$ (e.g., F1-measure).  
Given a fixed budget, we can ask questions
 until the budget is exhausted.
However, to achieve a  quality level $Q$, we must estimate the current error
of the trained classifier. 
The easiest way to do this is  to measure the trained classifier's ability to accurately classify the gold data according to
the desired quality metric.  We can then continue to ask questions until a
specific accuracy on the gold data is achieved (or until the rate of
improvement of accuracy levels off).

In the absence of (sufficient) gold data, we adopt the standard {\it
  $k$-fold cross validation} technique, randomly partitioning the
crowd-labeled data into test and training sets, and measuring the
ability of a model learned on training data to predict test values.
We repeat this procedure $k$ times and take the average as an overall assessment of the model's quality. 
Section~\ref{s:pba-expts} shows that this method
 provides more reliable estimates of the model's current quality
than relying on a small amount of gold data.

\subsection{Effect of Batch Sizes}
\label{sec:batch}

At each iteration of the iterative scenario, we must choose a subset of the unlabeled items according to their effectiveness scores, and send them to the crowd for labeling. 
We call this subset a ``batch'' (denoted as $U'$ in Line 5 of Figure~\ref{fig:iterative}). 
An interesting question is how to set this \emph{batch size}, say $\beta$.

Intuitively, a smaller $\beta$ increases opportunities to improve the AL algorithm's effectiveness by incorporating 
previously requested labels before deciding which labels to request next. 
For instance, best results are  achieved when $\beta$$=$$1$.
However, larger batch sizes reduce the overall run-time substantially by  
(i) allowing several workers to label items in parallel, and (ii) 
reducing the number of iterations.\footnote{As shown in Section~\ref{sec:scalability}, the time complexity of
each iteration is proportional to $T(|U|)$, where $T(.)$ is the training time
of the classification algorithm. For instance, consider the training of an SVM classifier which is typically
cubic in the size of its training set. Assuming a fixed $\beta$ and an overall budget of $B$ questions,
 the overall complexity of our \un algorithm becomes $O(\frac{B}{\beta}|U|^{3})$.} 
This is confirmed by our experiments in Section \ref{sec:optimization-expts}, which
 show that 
the impact of
 increasing $\beta$  on the effectiveness of our algorithms is not as dramatic as its impact on the overall run-time.
Thus, to find the optimal $\beta$, a reasonable choice is to start from a smaller batch size and continuously increase it (say, double it) until the run-time becomes reasonable, or the quality metric falls below the minimum requirement.

\if{0} 
 Alternatively, one could choose exponentially increasing batch-sizes, e.g. $a^{1}, a^{2}, \cdots, a^{t}$
 where $\sum_{i} a^{i}\leq B$, reducing the overall complexity to $O(log(B)|U|^{3})$ at the cost of degrading the algorithm's effectiveness
  at each iteration.
  This simple analysis clearly demonstrates the dramatic effect of batch-size on the overall run-time of our AL algorithm,
\fi

\if{0}


As mentioned in Section~\ref{sec:scalability} the run-time of each iteration of our \un algorithm is $O(m\cdot T(|U|))$ while it is
 $O((m+|U|)\cdot T(|U|))$ for \ete. Suppose we can label $B$ items in total, then a batch size of $b$ means we will run $\frac{B}{b}$ iterations.
Let $n$ and $n_{0}$ be the number of our initially labeled and unlabeled items, respectively. Also, let $T(n)$ and denote
the average running time of the training algorithm when the training size is $n$. Also,
let $C(n)$ be the average time needed for the crowd to label $n$ items (including the time needed for acquiring redundant labels for each item).

 Thus, the total time required by \un will be:
\begin{displaymath}
  f_{\un}(B,b) = \sum_{i=1}^{B/b} m\cdot T(n_{0}+i\cdot S) + C(b)
  \end{displaymath}

while the time of \ete will be:
\begin{displaymath}
  f_{\ete}(B,b) = \sum_{i=1}^{B/b} (m+(n-i\cdot bS))\cdot T(n_{0}+i\cdot b) + C(b)
  \end{displaymath}

{\bf Maximizing Model Quality Under Budget and Time Constraints.}

Our goal is to achieve the highest quality model, given a limited budget and time constraint.  We can formulate this as an optimization
problem. Suppose we have a labeling budget of $B$ items and a time constraint $t$, and that $t$ is large enough that we can at least
run our learner at the largest possible batch size (e.g., $B$).

will defini
Now the goal of our optimization for a given $B$ and $t$ is to maximize the quality $Q$, which is almost equivalent to minimizing $b$. In other words,

\begin{displaymath}
  \textrm{Minimize~} b \textrm{~subject to~} f(B,b) \leq t
  \end{displaymath}

 When the user only specifies $B$, we can choose the smallest \emph{feasible batch size} (i.e. a batch for which the crowd will still provide reasonable accuracy within reasonable time,
 in other words, if the batch size is too small most workers will be reluctant  or provide poor quality~\cite{?}).

When the user only specifies $t$, we can choose $B=n$ to allow for highest quality.

When only a quality measure is specified, we can start from the smallest feasible batch size and stop as soon as we reach Q to ensure minimizing the
cost.

-Say something about other cases

This analysis assumes that average estimations of $C(b)$ can be used. This itself is another research out of this paper's scope and has been
discussed in previous papers~\cite{?}, also assumes that a fixed batch is used throughout each iteration, and also assumes that the cost of asking
questions is the same for all questions.
\fi

\vspace*{-0.1in}
\section{Experimental Results}
\label{sec:expr}

This section evaluates the effectiveness of our AL algorithms in practice
by comparing the speed, cost, and accuracy with which our AL algorithms can label a dataset compared to
state-of-the-art AL algorithms.


\vspace{.1cm}
\noindent
\textbf{Overview of the Results.}
Overall, our experiments show the following: (i)  our AL algorithms 
require several orders of magnitude fewer questions to achieve the same quality
than the random baseline, and substantially fewer 
questions ($4.5\times$--$44\times$) than the best general-purpose AL algorithm (\iwal~\cite{para-active, iwal, iwalbb}),
(ii) our \ete algorithm works better than \un in the upfront setting, but the two are comparable in the iterative setting, (iii) \un has 
a much lower computational overhead than \ete, and 
(iv) surprisingly, even though our AL algorithms are generic and widely applicable, they still perform comparably to and sometimes much better than AL algorithms designed for specific tasks, e.g., $7\times$  fewer questions than 
\crowdER~\cite{crowd-er}  and  an order of magnitude fewer than \aditya~\cite{active-sampling}
(two of the most recent AL algorithms for entity resolution), competitive results to Brew et al \cite{al-sentiment},
and also
$2$--$8\times$ fewer questions than less general AL algorithms (\pst~\cite{provost} and \smd~\cite{svm-margin}).
 
\vspace{.1cm}
\noindent
\textbf{Experimental Setup.} All algorithms were tested on a Linux server with
 dual-quad core Intel Xeon 2.4 GHz processors and 24GB of RAM. Throughout this section, unless stated otherwise, we repeated each experiment   $20$ times and reported the average result, every task cost 1\textcent, and the size of the initial training and the batch size were  $0.03\verb=%=$ and $10\verb=%=$ of 
 the unlabeled set, respectively.

\vspace{.1in}
\noindent{\bf Methods Compared.} We ran experiments on the following learning algorithms in both the upfront and iterative scenarios:
\begin{asparaenum}
\item {\it \un}:  Our method from Section~\ref{sec:uncertainty}.
\item {\it \ete}: Our method from Section~\ref{sec:ete}.
\item {\it \iwal}: A popular AL algorithm \cite{iwalbb} that follows Importance Weighted Active Learning \cite{iwal},
 recently extended with batching \cite{para-active}.
\item {\it \pst}: Another bootstrap-based AL that uses the model's class probability estimates to measure uncertainty~\cite{provost}. This method only works for probabilistic classifiers, e.g., we exclude this in experiments with SVMs.
\item {\it \crowdER}: One of the most recent AL techniques specifically designed for entity resolution tasks~\cite{crowd-er}. 
\item {\it \aditya}: Another state-of-the-art AL specifically designed for entity resolution~\cite{active-sampling}.
%

\item {\it \smd}: An AL algorithm specifically designed for SVM classifiers, which picks items that are closer to the margin~\cite{svm-margin}.

\item {\it \ent}: A common AL strategy \cite{al-survey} that picks items for which  the entropy of different class probabilities is higher, i.e., 
the more uncertain the classifier, 
the more similar the probabilities of different classes, and the higher the entropy.

\item {\it Brew et al. \cite{al-sentiment}}: a domain-specific AL designed for sentiment analysis, which 
uses clustering  to select an appropriate subset of articles (or tweets)  
 to be tagged by users.

\item {\it Baseline}: A passive learner that randomly selects unlabeled items to send to the crowd.
\end{asparaenum}
In the plots, we prepend the scenario name to the algorithm names, e.g., Upfront\ete or IterativeBaseline.
We have repeated our experiments with different classifiers as the underlying learner, including SVM, Na\"{\i}ve-Bayes classifier,
neural networks, and decision trees. For lack of space, we only report each experiment for one type of classifier. When not specified, we used linear SVM.

\label{sec:metrics}
\vspace{.1in}
\noindent{\bf Evaluation Metrics.} AL algorithms are usually evaluated based on their \emph{learning curve}, which plots the quality measure of interest (e.g., accuracy or F1-measure) as a function of the number of data items that are labeled~\cite{al-survey}.
To  compare different learning curves quantitatively, the following metrics are typically used:

\begin{asparaenum}

\item Area under curve (AUC) of the learning curve.


\item AUCLOG, which is the AUC of the learning curve when the X-axis is in log-scale.


\item Questions saved, which is the ratio of number of questions asked by an
active learner to those asked by the baseline to achieve the same quality.

\end{asparaenum}

Higher AUCs indicate that the learner achieves a higher quality for the same cost/number of questions.
Due to the diminishing-return of learning curves, the average quality improvement is usually in a $0$--$16\verb=%=$ range.

AUCLOG favors algorithms that improve the metric of interest early on (e.g., with few examples). Due to the logarithm,
the improvement of this measure is typically in the $0$--$6\verb=%=$ range.

To compute question  savings, we average over all the quality levels that are achievable by both AL and baseline curves. 
 For competent active learners, this measure should (greatly) exceed $1$, as
a ratio $<1$ indicates a performance worse than that of the random baseline.




\subsection{Crowd-sourced Datasets}
\label{sec:domains}

We experiment with several datasets labeled using Amazon Mechanical Turk. In this section, we report the performance
of our algorithms on each of them.

\vspace{-0.1cm}
\subsubsection{Entity Resolution}
\label{sec:entity}

Entity resolution (ER) involves finding different records that refer to the same entity, and is an essential step in data integration/cleaning \cite{arasu-er,active-sampling,Qurk,sarawagi-er,crowd-er}.
Humans are typically more accurate at ER than classifiers, but also slower and more expensive~\cite{crowd-er}.

\begin{figure}
\centering
\vspace{0.1in}
\includegraphics[width=2.4in]{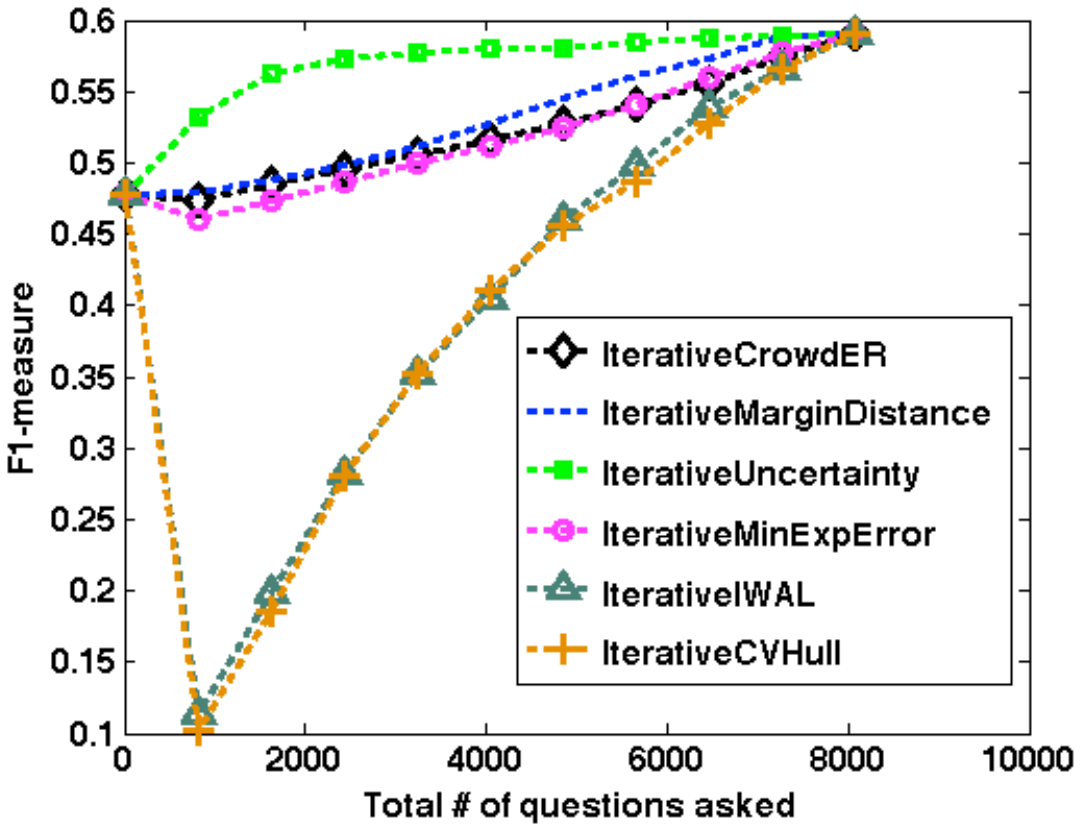}
\caption{{\small Comparison of different AL algorithms for entity resolution: the overall F1-measure in the iterative scenario.}}
	\label{fig:big-entity-overall}
\end{figure}

We used the Product ({\scriptsize \url{http://dbs.uni-leipzig.de/file/Abt-Buy.zip}}) dataset,
which contains product attributes (name, description, and price) of items listed on the \url{abt.com} and \url{buy.com} websites.
The task is to detect pairs of items that are identical but listed under different descriptions on the two websites
 (e.g., ``iPhone White 16 GB'' vs ``Apple 16GB White iPhone 4'').
 We used the same dataset  as \cite{crowd-er}, where the crowd was 
asked to label $8315$ pairs of items as either identical or non-identical.
This dataset consists of $12\verb=%=$ identical pairs and $88\verb=%=$ non-identical pairs.
In this dataset, each pair has been labeled by $3$ different workers, with an average accuracy of $89\verb=%=$ and an F1-measure of $56\verb=%=$.
We also used the same classifier used in~\cite{crowd-er}, namely a linear SVM where each pair of items is represented by their Levenshtein and Cosine similarities.
When trained on $3\verb=%=$ of the data, this classifier has an average accuracy of $80\verb=%=$ and an F1-measure
 of $40\verb=%=$. 
 Figure \ref{fig:big-entity-overall} shows the results of using different AL algorithms. 
  As expected, while all methods eventually improve with more questions,
  their overall F1-measures 
  improve at different rates. \smd, \ete, and \crowdER are all comparable, while
 \un improves much 
 more quickly than the others. Here, \un can  identify the items about which the model has the most uncertainty and get the crowd to label those earlier on. 
 Interestingly, \iwal, which is a generic state-of-the-art AL, performs extremely poorly in practice.
\aditya performs equally poorly, as it internally relies on \iwal as its AL subroutine. 
This suggests opportunities for extending  \aditya to rely on other AL algorithms in future work.

 This result is highly encouraging: even though \crowdER and \aditya are recent AL algorithms highly \emph{specialized} 
 for improving the recall (and indirectly, F1-measure) of ER, our \emph{general-purpose} AL algorithms are still 
 quite competitive. In fact, \un uses $6.6\times$ fewer questions than \crowdER and an 
 order of magnitude fewer questions than \aditya to achieve the same F1-measure.

\subsubsection{Image Search}
\label{sec:vision}

Vision-related problems also utilize crowd-sourcing
heavily, e.g., in tagging pictures, finding objects, and identifying bounding boxes~\cite{vision-al}. In all of our vision experiments, we 
employed a relatively simple classifier where the PHOW features (a variant of dense SIFT descriptors commonly used in vision tasks~\cite{vrClassify}) of a set of images are first extracted as a bag of words, and then a linear 
SVM is used for their classification. 
Even though this is not the state-of-the-art image detection algorithm, we show that our AL algorithms still 
greatly reduce the cost of many challenging vision tasks.

\begin{figure*}[th]
\centering
\includegraphics[width=6.8in]{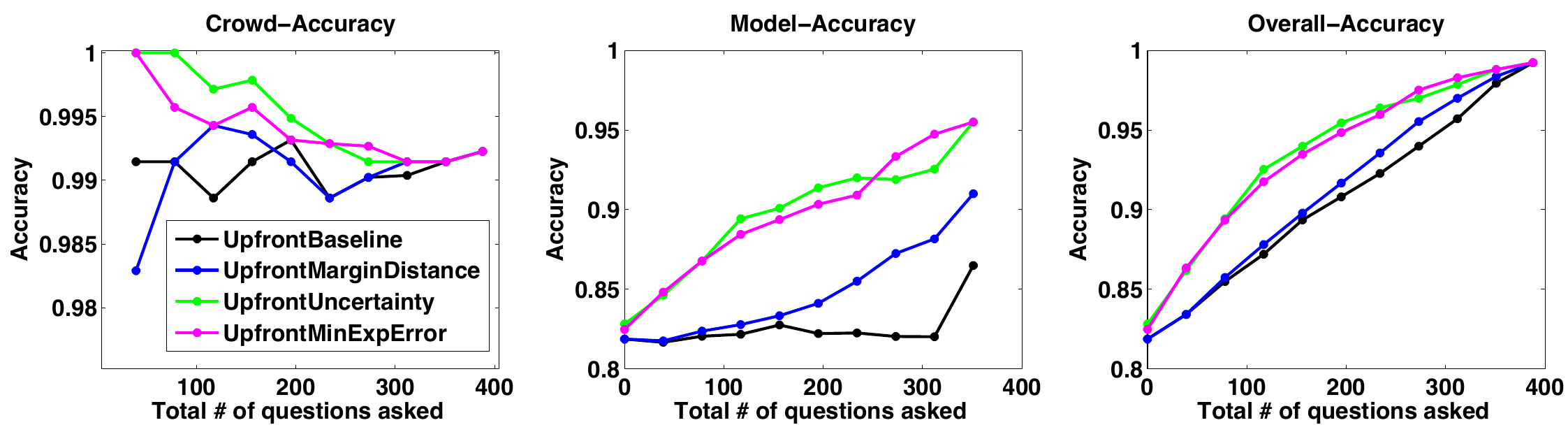}
\caption{\small{The object detection task (detecting the gender of  the person in an image): accuracy of the (a) crowd, (b) model, (c) overall.}}
	\label{fig:gender-accuracy}
\vspace{0.2in}	
\end{figure*}

\begin{figure}
\centering
\includegraphics[width=2.4in]{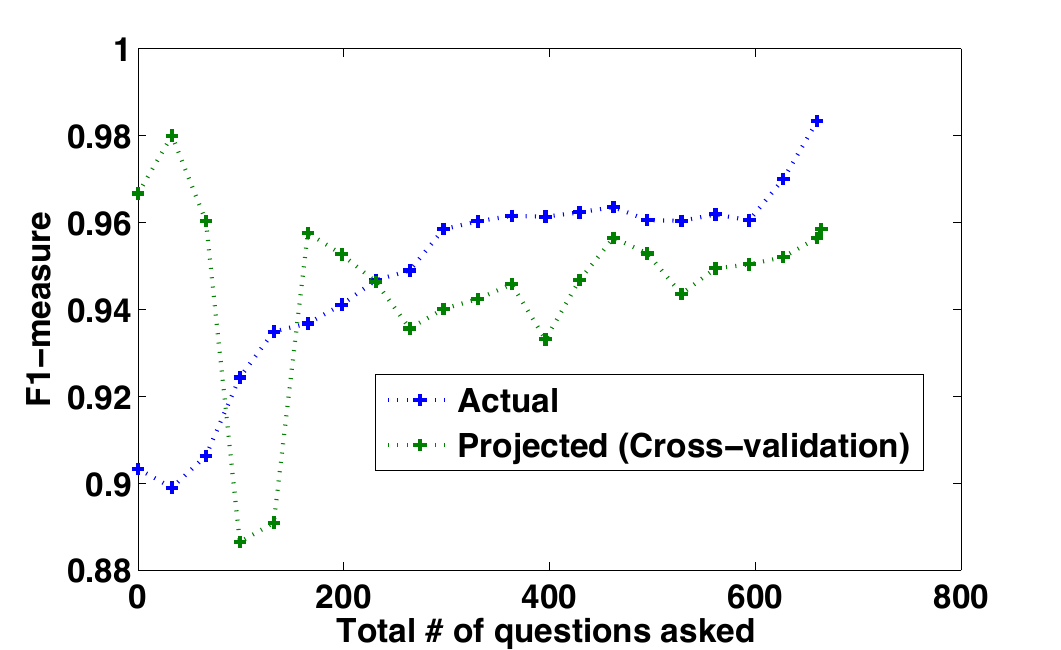}
\vspace{-0in}
\hspace{-.2in}
\caption{{\small Estimating quality with $k$-fold cross validation.}}
\label{fig:cross-validation}
\end{figure}

\noindent\textbf{Gender Detection.} We used the faces from Caltech101 dataset~\cite{caltech101} and manually labeled each image with its
gender (266 males, 169 females) as our ground truth. We also gathered crowd labels by asking the gender of each image from $5$ different workers. We started by training the model on  a random set of $11\verb=%=$ of the data.
In Figure~\ref{fig:gender-accuracy}, we show the accuracy of the crowd, the accuracy of our machine
 learning model, and the overall accuracy of the model plus crowd data. For instance, when a fraction  
$x$  of the labels were obtained from the crowd, the other $1-x$ labels were determined from the model, and thus, the overall accuracy was  $x*a_{c}+(1-x)*a_{m}$, where $a_{c}$ and $a_{m}$ 
are the crowd and model's accuracy, respectively.
As in our entity resolution experiments, our algorithms  improve 
the quality of the labels provided by the crowd, i.e., by asking questions for which the crowd tends to be more reliable. 
Here, though, the crowd produces higher overall quality than in the entity resolution case and therefore its accuracy is 
improved only from $98.5$\verb=%= to $100\verb=%=$.
Figure~\ref{fig:gender-accuracy} shows
that both \ete and \un perform well in the upfront scenario, 
respectively improving the baseline accuracy 
by $4\verb=%=$ and $2\verb=%=$ on average, and improving its AUCLOG by $2$-$3\verb=%=$.  Here, due to the upfront scenario, \ete saves 
the most number of questions.
The baseline has to ask $4.7\times$ ($3.7\times$) more questions than \ete (\un) to achieve the same accuracy.
Again \smd, although specifically designed for SVM, achieves little improvement over the baseline.

\begin{figure*}[th]
\begin{minipage}{2.2in}
\centering
\includegraphics[width=2.1in]{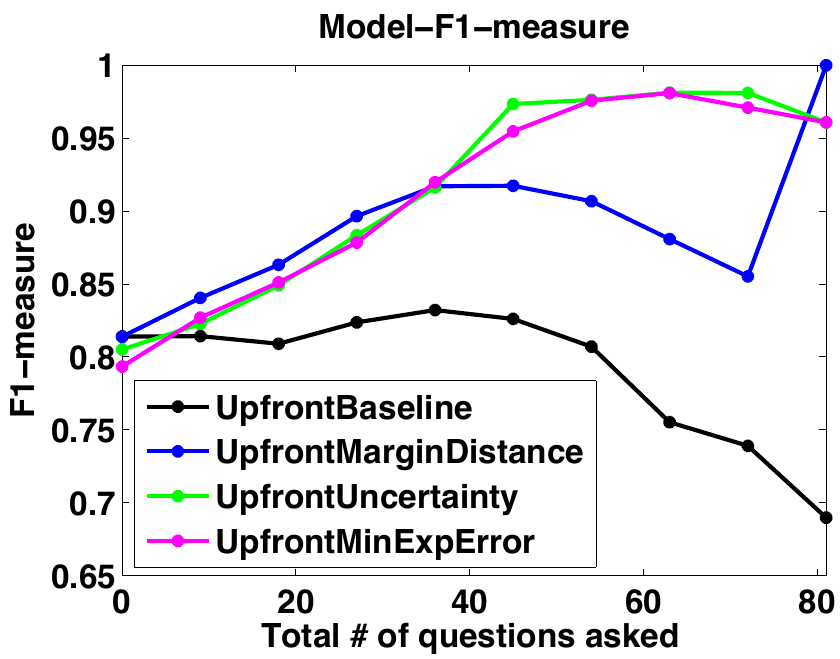}
\caption{{\small Image search task (whether a scene contains a human): F1-measure of the model.}}
	\label{humanface-fmeasure}
\end{minipage}
\hspace{.2in}
\begin{minipage}{2.2in}
\includegraphics[width=2.1in]{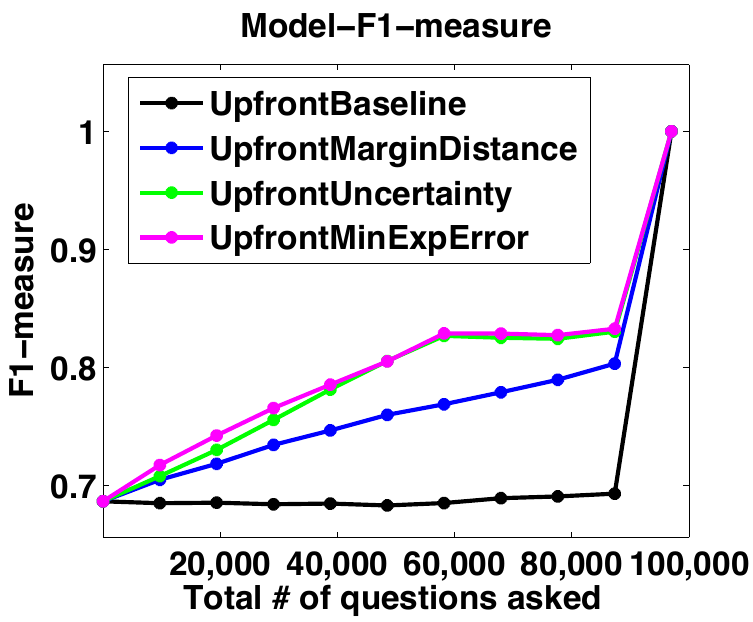}
\caption{{\small Sentiment analysis task: F1-measure of the model for $100$K tweets.}}
	\label{fig:tweets-fmeasures}
\end{minipage}
\hspace{.2in}
\begin{minipage}{2.2in}
\includegraphics[width=2.1in]{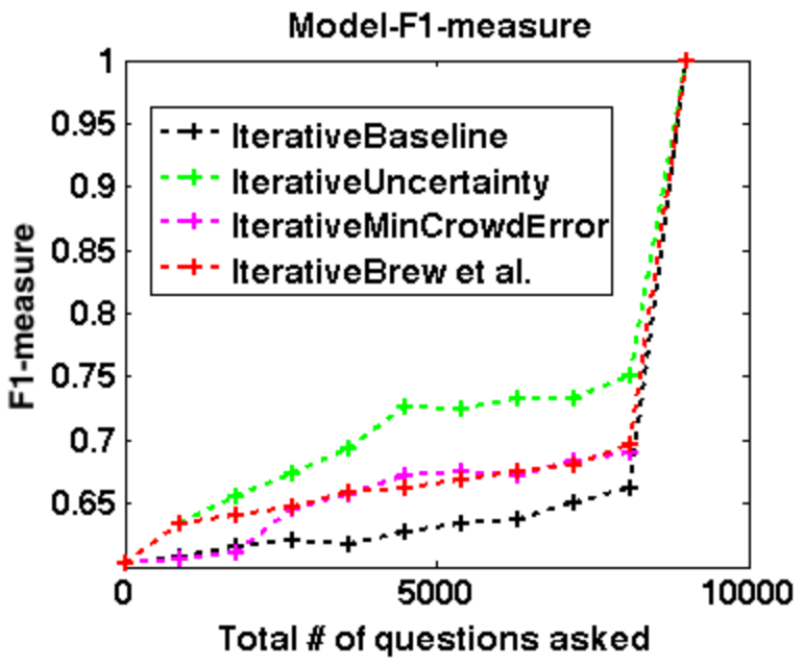}
\caption{{\small Sentiment analysis task: F1-measure of the model for $10$K tweets.}}
	\label{fig:tweets-iter-fmeasure-brew}
\end{minipage}
\vspace{0.2in}
\end{figure*}

\noindent\textbf{Object Containment.} We again mixed $50$ human faces and $50$ background images from Caltech101~\cite{caltech101}.  Because
differentiating human faces from background clutter is easy for humans,  we used the crowd labels as ground truth in this experiment. 
Figure~\ref{humanface-fmeasure} shows the upfront scenario with an initial set of $10$ labeled images, where both \un and \ete lift the baseline's F1-measure by $16\verb=%=$, while \smd provides a lift of 13\verb=%=. All three algorithms increase the baseline's AUCLOG by $5$-$6\verb=%=$. Note that the baseline's 
F1-measure degrades slightly as it reaches higher budgets, 
since the baseline is forced to give answers to hard-to-classify questions, 
while the AL algorithms avoid such questions, leaving 
them to the last batch (which is answered by the crowd). 

\eat{
\begin{figure}[th]
\centering
\includegraphics[height=4.8cm]{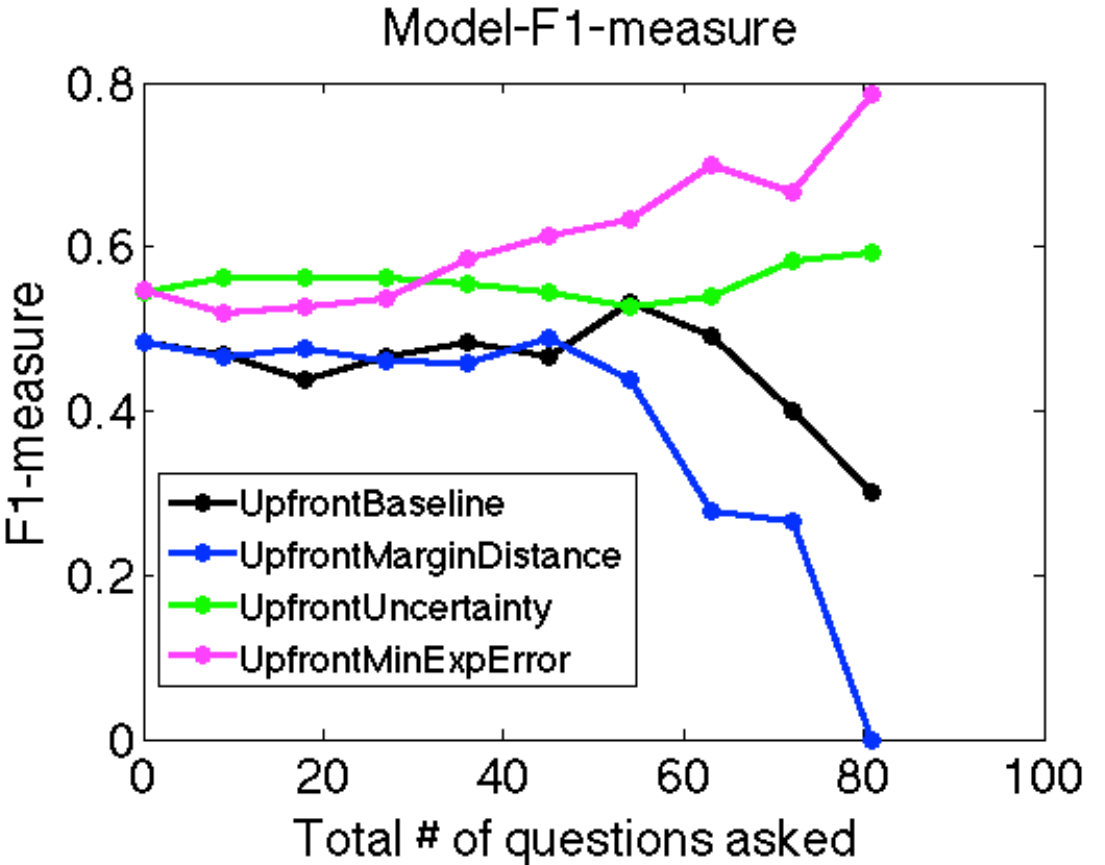}
\caption{The objection categorization task (separating crayfish from lobster pictures). F1-measure of the model.}
	\label{lobster-fmeasure}
\end{figure}

In Figure~\ref{lobster-fmeasure}, both \ete and \un provide significant improvements: 40\verb=%= and 26\verb=%=, reps with the former being
more effective due to the upfront scenario. They both provide about 19\verb=%= lift in AUCLOG. MarginDistance performs very poorly.
}

\subsubsection{Sentiment Analysis}
\label{sec:twitter}
Microblogging sites such as Twitter provide rich datasets for  
sentiment analysis~\cite{sentiment-twitter},
where analysts can ask questions such as ``how many of the tweets mentioning iPhone have a positive or negative sentiment?''
Training accurate classifiers requires sufficient accurately labeled data, and
with millions of daily tweets, it is too costly to ask the crowd to label all of them.
In this experiment, we show that our AL algorithms,  
with as few as $1$K-$3$K crowd-labeled tweets,  can achieve very high accuracy and F1-measure on 
 a corpus of $10$K-$100$K unlabeled tweets.
We randomly chose these tweets from an online corpus\footnote{http://twittersentiment.appspot.com} that provides ground truth labels for the tweets,
with equal numbers of positive and negative-sentiment tweets. We obtained crowd labels ({\it positive}, {\it 
negative}, {\it neutral}, or {\it vague/unknown}) for each tweet from $3$ different workers.

Figure~\ref{fig:tweets-fmeasures} shows the results for using $1$K initially labeled data points with the $100$K dataset in the upfront setting.
 The results confirm that the upfront scenario is best handled by our \ete Algorithm. Here, the \ete, \un and \smd algorithms improve the average F1-measure of the
 baseline model by
$11\verb=%=$, $9\verb=%=$ and $5\verb=%=$, respectively.
Also, \ete increases baseline's AUCLOG by $4\verb=%=$. All three AL algorithms dramatically reduce the number of  questions required to achieve a given accuracy or F1-measure.  In comparison
to the baseline, \ete, \un, and \smd
reduce the number of questions by factors of $46\times$, $32\times$, and $27\times$, respectively.

Figure~\ref{fig:tweets-iter-fmeasure-brew} shows similar results for using $1$K initially labeled tweets with the $10$K dataset but in the iterative setting.
 The results confirm that the iterative scenario is best handled by our \un algorithm, which even 
 improves on Brew et al. algorithm \cite{al-sentiment}, which is a domain-specific AL designed for sentiment analysis. Here, the \un, \ete, and Brew et al. algorithms improve the average F1-measure of the
 baseline model by
$9\verb=%=$, $4\verb=%=$ and $4\verb=%=$, respectively.
Also, \un increases baseline's AUCLOG by $4\verb=%=$. 
In comparison
to the baseline, \un, \ete, and Brew et al.
reduce the number of questions by factors of $5.38\times$, $1.76\times$, and $4.87\times$, respectively.
Again, the savings are expectedly modest compared to the upfront scenario.

\if{0}
Given the smaller size of  the $10$K dataset, however, the iterative scenario seems more feasible. Results of running iterative are shown in
 Figure~\ref{fig:tweets-fmeasures} (b), where as expected, the best improvement is achieved by our \un Algorithm.
Here, the baseline's F1-measure is on average lifted by $12\verb=%=$ using \un while \ete and \smd only slightly improve on the baseline, i.e. $1$-$2\verb=%=$.
Moreover, \ete increases the AUCLOG by 2\verb=%= and reduces the number of questions for the same \emph{overall F1-measure}
by 33\verb=%=.
In terms of reducing the number of questions for the achieving the same F1-measure, the baseline needs  $4.6\times$,  $1.8\times$ and
$1.1\times$ more questions than \un, \ete, and \smd, respectively.
\fi

\subsection{\pba~and other Crowd Optimizations}
\label{s:pba-expts}
\label{sec:optimization-expts}

In this section, we present results for our crowd-specific optimizations described in Section~\ref{sec:optimization}.

\begin{figure}[t]
\centering
\includegraphics[height=1.6in, width=1.8in]{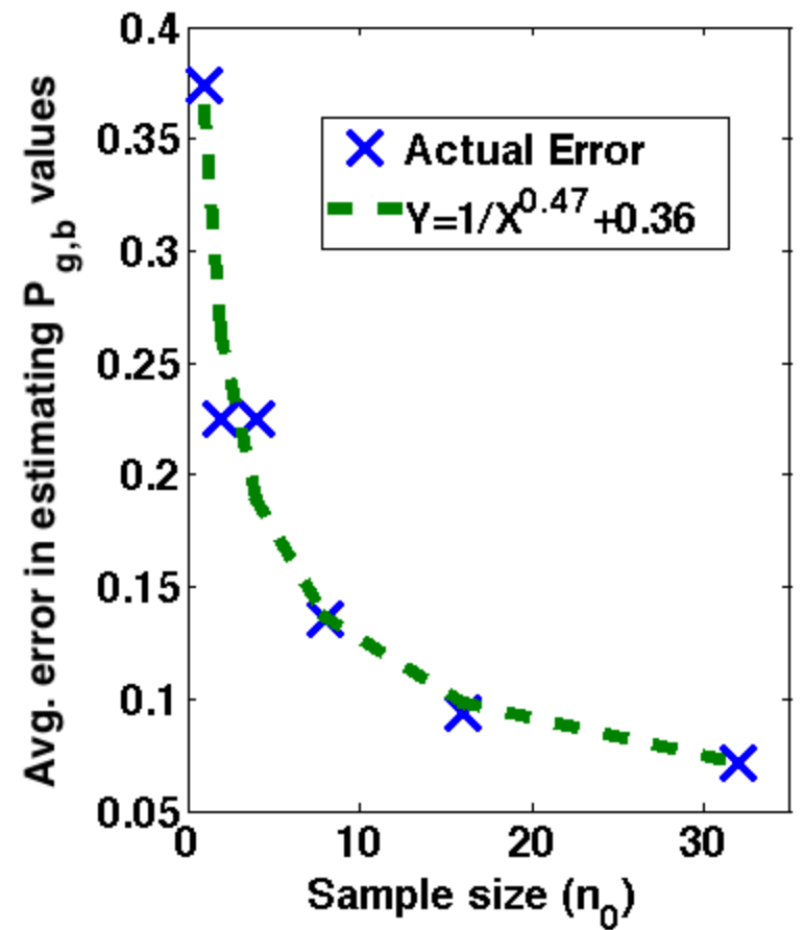}
\caption{{\small The effect of sample size on the accuracy of $P_{g,b}$ estimates in the \pba algorithm.}}
\label{fig:pba-error-estimation}
\end{figure}

\vspace{.1cm}
\noindent {\bf PBA Algorithm:}
We first report experiments on the {\it PBA} algorithm.
Recall that this algorithm partitions the items into subgroups
and optimally allocates the budget amongst them.
In the CMU facial expressions dataset, the crowd had a particularly hard time
telling the facial expression of certain individuals, 
so we created subgroups based on the
{\it user} column of the dataset, and asked the crowd to label the expression on each face.
By choosing $v_{0}$=$9$ and $b^{max}$=$9$, we 
varied $n_{0}=2,4,8,16,32$ and measured the point-wise error of our  $P_{g,b}$  estimates. 
Figure \ref{fig:pba-error-estimation} shows that 
the average error of our estimates drops quickly as the sample size $n_0$ increases: 
the error is $0.22$ for $n_0$=$2$ and goes down to $0.13$ with only $n_0$=$8$.
Also,  curve fitting shows that the error is proportional to $O(1/n_0)$ (like most aggregates).

\begin{figure}[t]
\centering
\includegraphics[width=2.4in]{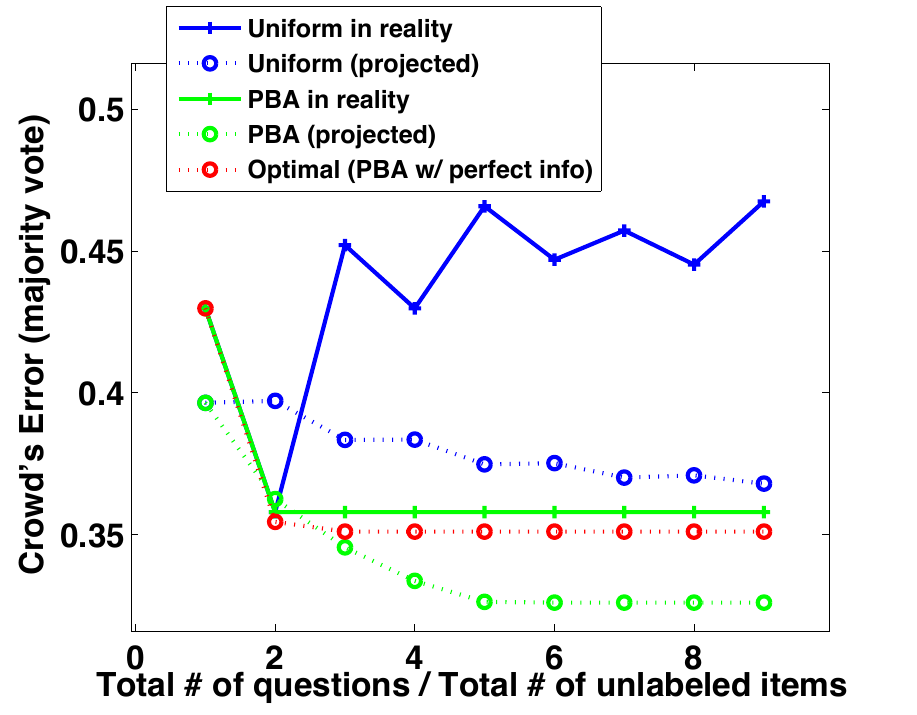}
\caption{{\small Reducing the crowd's noise under different budget allocation schemes.}}
\label{fig:optimal-allocation}
\end{figure}

Choosing $v_{0}=9$, $b^{max}=9$, and
$n_0=2$,  we also compared {\it PBA} against
a uniform budget allocation 
scheme, where the same number of questions are asked about all items uniformly, as done in previous research
 (see~\ref{sec:related}).
 The results are shown in Figure~\ref{fig:optimal-allocation}.  Here, the X axis shows the normalized budget, e.g., a value of $2$ means 
 the budget was twice the total number of unlabeled items. 
  The  Y axis shows the overall (classification) error of the crowd using majority voting under different allocations.
 Here, the solid lines show the actual error achieved under both strategies, while the blue and green dotted lines show
  our estimates of their performance before running the algorithms.
       From Figure~\ref{fig:optimal-allocation},  we see that although our estimates of the actual error are not highly accurate,
       since we only use them to solve an ILP that would favor harder subgroups,
       our {\it PBA} algorithm (solid green) still reduces
       the overall crowd error by about $10\verb=%=$ (from $45\verb=%=$ to $35\verb=%=$).
        We also show how {\it PBA} would perform if it had an oracle that provided access to exact values of
   $P_{g,b}$ (red line).

\begin{figure*}
\hspace{.15in}
\begin{minipage}{2in}
\centering
\includegraphics[height=1.8in]{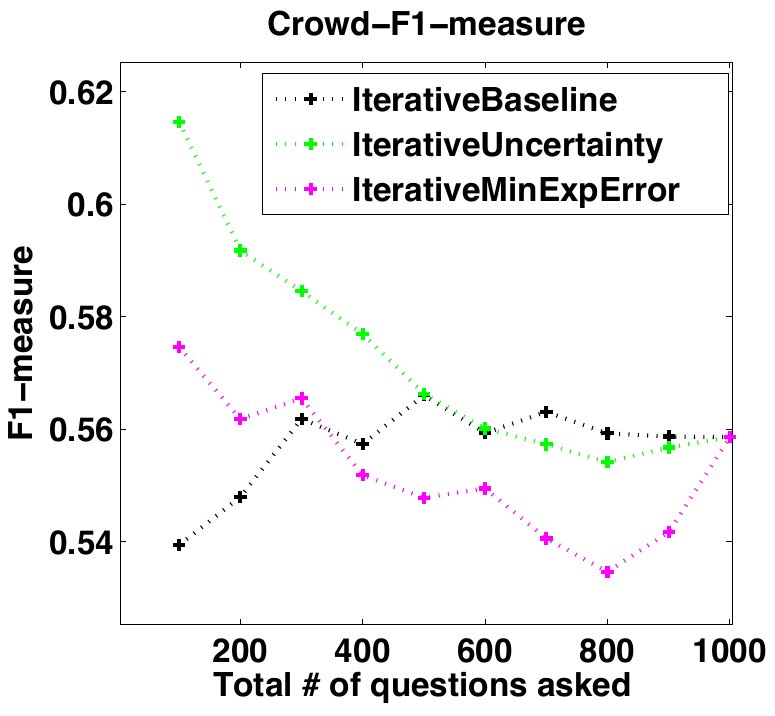}
\vspace{-.15in}
\caption{{\small Improving the crowd's F1-measure for entity resolution in the iterative scenario.}}
	\label{fig:small-entity-crowd}
\end{minipage}
\hspace{.6in}
\begin{minipage}{1.7in}
\hspace{-.5in}
\centering\includegraphics[height=1.7in]{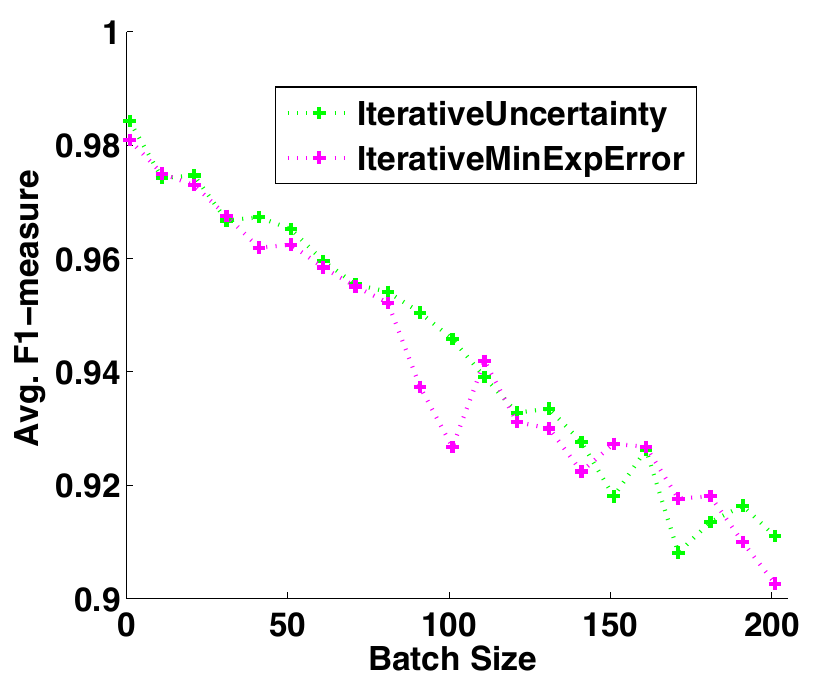}
\vspace{-.15in}
\caption{{\small Effect of batch size on our algorithms' F1-measure (vehicle dataset, w/ a budget of 400 questions).}}
\label{fig:batchsize-quality}
\end{minipage}
\hspace{.6in}
\begin{minipage}{1.6in}
\centering
\hspace{-.6in}
\includegraphics[height=1.7in]{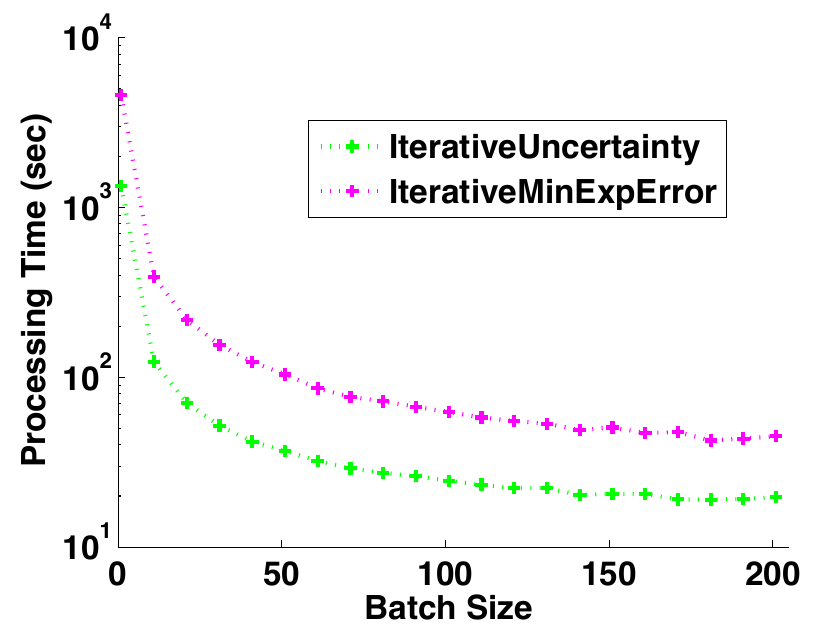}
\vspace{-.25in}
\caption{{\small Effect of batch size on our algorithms'  processing times (vehicle dataset, w/ a budget of 400 questions).}}
	\label{fig:batchsize-runtime}
\end{minipage}
\end{figure*}

\begin{figure}[t]
\centering
\includegraphics[width=2in, height=1.5in]{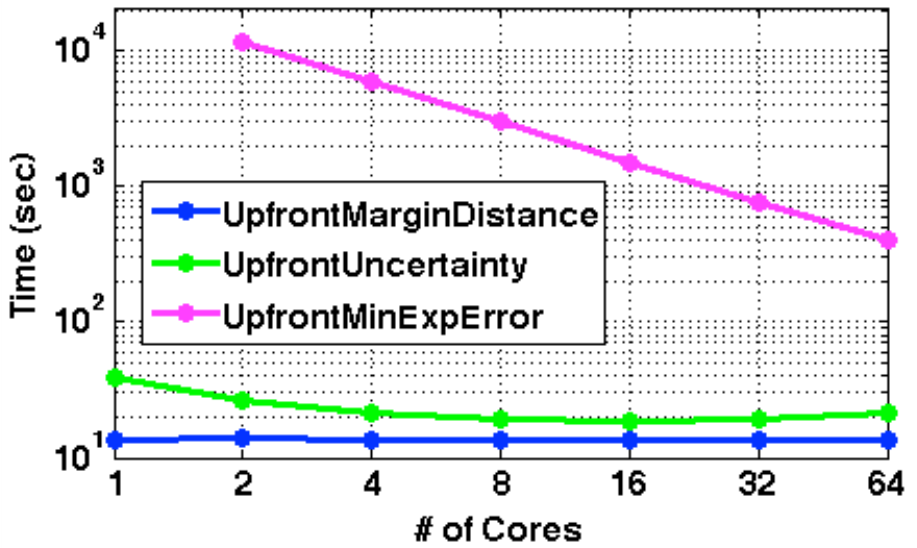}
\vspace{-.1in}
\caption{{\small Effect of parallelism on processing time: 100K tweets.}}
	\label{fig:scale}
\end{figure}

\vspace{.1cm}
\noindent {\bf Balancing Classes:}
Recall from Section~\ref{sec:noise} that our algorithms tend to ask more questions about rare classes than 
common classes, which can improve the overall performance.
Figure~\ref{fig:small-entity-crowd} reports the crowd's F1-measure
in an entity resolution task under different AL algorithms.
The dataset used in this experiment has many fewer positive instances ($11\verb=%=$) than negative ones ($89\verb=%=$).
 The main observation here is that although the crowd's average F1-measure for the entire dataset is $56\verb=%=$
(this is achieved by the baseline),
our \un algorithm can lift this up to $62\verb=%=$, mainly because the  questions posed to the crowd have a
 balanced mixture of positive and negative labels. 

\vspace{.1cm}
\noindent {\bf k-Fold Cross Validation for Estimating Accuracy: }
We use k-fold cross validation to estimate the current quality of our model.
Figure~\ref{fig:cross-validation} shows our estimated F1-measure for an SVM classifier
on UCI's cancer dataset. Our estimates are reasonably close to the true F1 values, especially as more labels are obtained from the crowd.
This suggests that k-fold cross validation allows us to effectively estimate current model accuracy, and to stop
acquiring more data once model accuracy has reached a reasonable level.

\vspace{.1cm}
\noindent {\bf The Effect of Batch Size:}
We now study the effect of batch size on result quality, based on the observations in
Section~\ref{sec:batch}.
The effect is typically moderate (and often linear), as shown
in Figure~\ref{fig:batchsize-quality}.  Here we show that the F1-measure gains can be
in the 8--10\verb=%= range (see Section~\ref{sec:batch}).
However, larger batch sizes reduce runtime substantially, as Figure \ref{fig:batchsize-runtime} shows.  Here, going from
batch size $1$ to $200$ significantly reduces the time to train a model, by about two orders of magnitude (from
$1000$'s of seconds to $10$'s of seconds). 

\vspace{-0.1cm}
\subsection{UCI Classification Datasets}
\label{sec:uci}

\begin{figure}[tb]
\vspace{0.1in}
\centering\includegraphics[height=4.1cm]{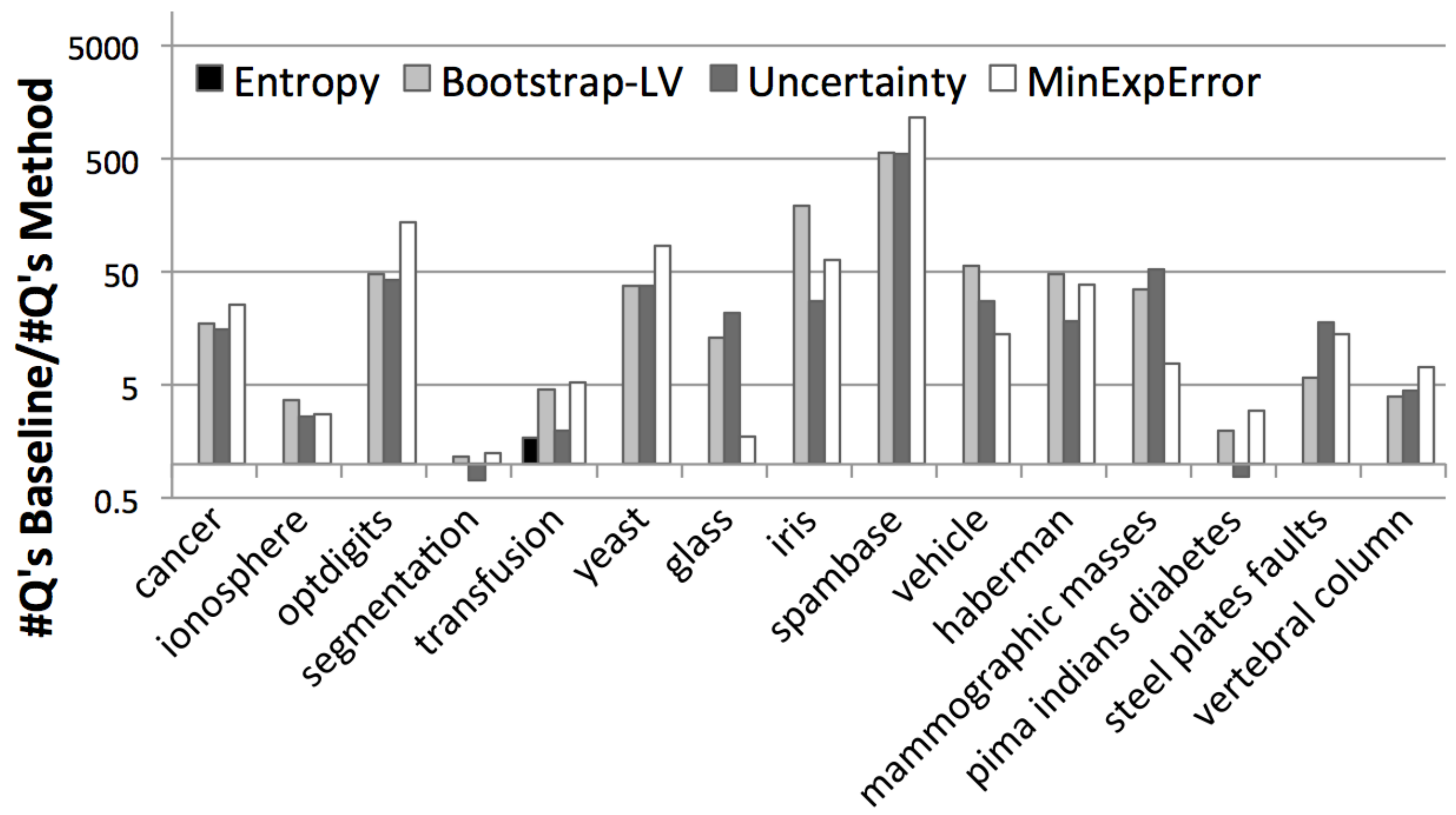}
\caption{{\small 
The ratio of the num. of questions asked by the random baseline to those asked by different AL algorithms in the Upfront scenario.
}}
\label{upfront-bars}
\end{figure}

\begin{figure}[tb]
\vspace{0.1in}
\centering\includegraphics[height=4cm]{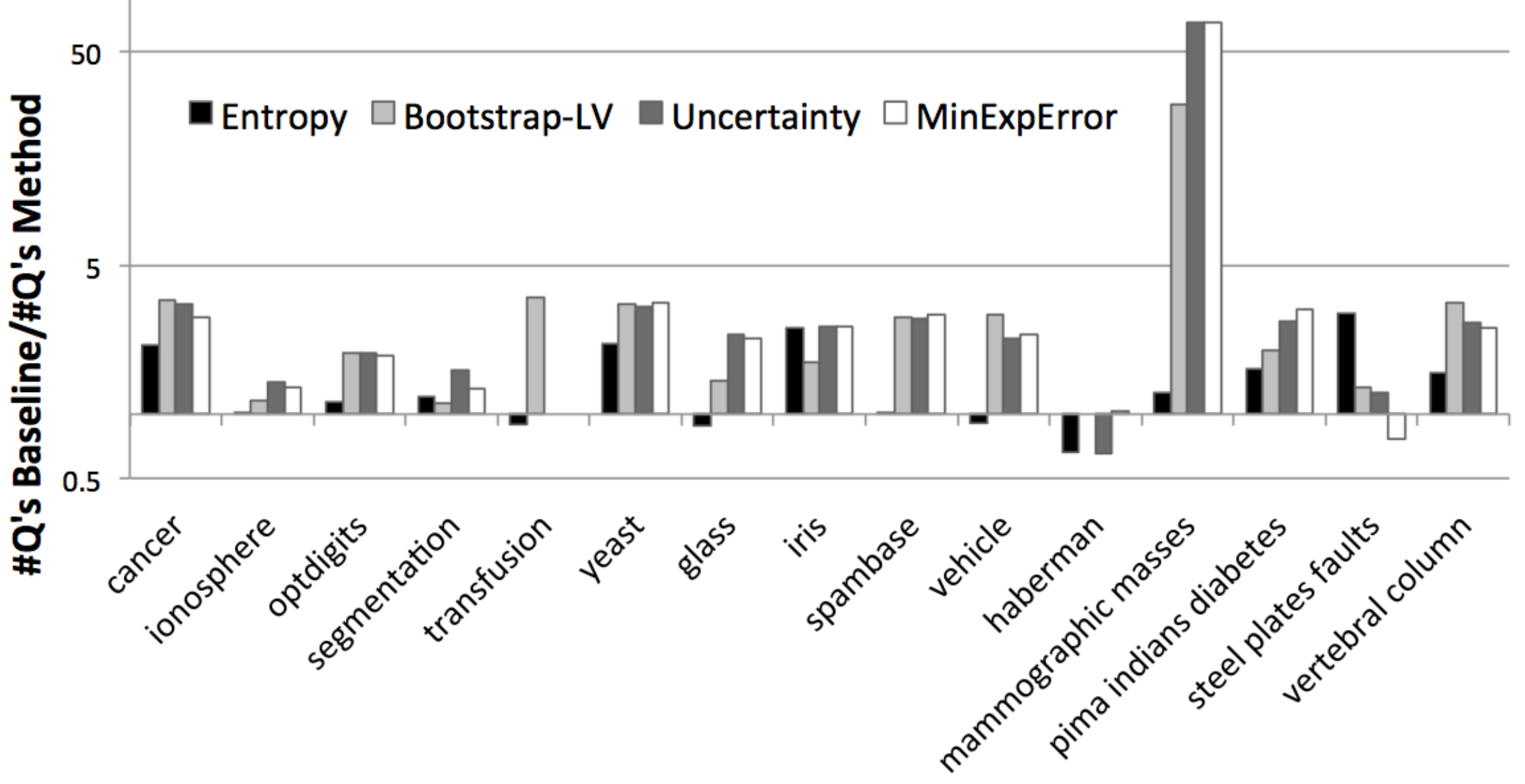}
\caption{{\small The ratio of the num. of questions asked by the random baseline to those asked by different AL algorithms in the Iterative scenario.}}
\label{iterative-bars}
\end{figure}

In Section~\ref{sec:domains}, we validated our algorithms on crowd-sourced datasets. This section  also compares our algorithms on datasets from the UCI KDD \cite{uci-kdd}, where labels are provided by
experts;  that is, ground truth and crowd labels are the same. Thus, by excluding 
the effect of noisy labels, we can
compare different AL strategies in isolation. We have chosen  $15$ well-known datasets, as shown in
Figures~\ref{upfront-bars} and~\ref{iterative-bars}.
To avoid bias, we have avoided any dataset-specific tuning or preprocessing steps, and applied the same classifier with the same settings
to all datasets.
In each case, we experimented with $10$ different budgets of
$10\verb=%=, 20\verb=%=, \cdots, 100\verb=%=$ (of total number of labels), each repeated $10$ times, 
and reported the average. 
Also, to compute the F1-measure for datasets with more than $2$ classes,
we have either grouped all non-majority classes into a single class, or arbitrarily partitioned all the classes into two new ones.

 Here, besides the random baseline, we compare \un and \ete against four other AL techniques, namely
 \iwal, \\
 \smd, \pst, and \ent. \iwal is as general as our algorithms, \smd only applies to SVM classification, and
\pst and \ent are only applicable to probabilistic classifiers.
For all methods (except for \smd) we used MATLAB's decision trees as the classifier, with its
default parameters except for the following: no pruning,
no leaf merging, and a `minparent' of $1$ (impure nodes with $1$ or more items can be split).

Figures~\ref{upfront-bars} and~\ref{iterative-bars} show the reduction in the number of questions
 under both upfront and iterative settings for \ent, \pst, \un, and \ete, while Table~\ref{tab:avg-improvement}
shows the average AUCLOG, F1, and reduction in the number of questions asked across all 15 datasets
for all AL methods.  The two figures omit detailed results for \smd and \iwal, as they performed poorly 
(as indicated in Table~\ref{tab:avg-improvement}).
We report all the measures of different AL algorithms in terms of their performance improvement relative to the baseline (so  higher numbers are better).
For instance, consider Figure~\ref{upfront-bars}. On the yeast dataset, \ete reduces the number of questions asked by $84\times$, while
 \un and \pst reduce it by about $38\times$ and \ent does not improve the baseline.

In summary, these results are consistent with those observed with crowd-sourced datasets. In the upfront setting, \ete significantly outperforms other AL techniques, with more than $104\times$ savings in the total number of questions on average.
\ete also improves the AUCLOG and average F1-measure of the baseline on average by $5\verb=%=$ and $15\verb=%=$, respectively.
 After \ete, the \un and \pst are most effective with a comparable performance, i.e. $55$-$69\times$ savings, improving the AUCLOG by $3\verb=%=$,
  and lifting the average F1-measure by
   $11$-$12\verb=%=$. 
 \pst  performs well here, 
which we expect is due to its use of bootstrap (similar to our algorithms). 
However, 
recall that \pst only works for probabilistic classifiers (e.g., decision trees).
Here, \smd is only moderately effective, providing around $13\times$ savings.
Finally, the least effective algorithms are \iwal and \ent, which  perform quite poorly across almost all datasets. 
\iwal uses learning theory to establish worst-case bounds on sample complexity (based on VC-dimensions), but
 these bounds are known to leave a significant gap between theory and practice, as seen here. 
\ent relies on the classifier's own class probability estimates \cite{al-survey}, and thus can be quite ineffective 
when these estimates are highly inaccurate. To confirm this, we used bootstrap to estimate the class probabilities more accurately 
(similarly to our \un algorithm), and then computed the entropy of these estimates. The modified version, denoted as \un(\ent),  
is significantly more effective than the baseline ($73\times$), which shows that our idea of using bootstrap in AL not only 
achieves generality (beyond probabilistic classifiers) but can also improve traditional AL strategies by providing more accurate probability estimates.

For the iterative scenario, \un actually works  better than \ete, with an average saving of $7\times$ over the baseline in questions asked and an increase in
AUCLOG and average F1-measure by $1\verb=%=$ and $3\verb=%=$, respectively. 
Note that savings are generally more modest than in the upfront case because the baseline receives much more labeled data in the iterative setting and
therefore, its average performance is much higher, leaving less room for improvement. 
However, given the comparable (and even slightly better) performance of \un compared to \ete in the iterative scenario, 
\un becomes a preferable choice for this scenario due to its considerably smaller  processing overhead (see Section \ref{sec:exp:scalability}).

\begin{table}[htb]
\small
\begin{tabular}{|c|c|c|c|}
\hline
\multicolumn{4}{|c|}{\bf Upfront} \\
\hline
  Method & AUCLOG(F1) & Avg(Q's Saved) & Avg(F1) \\
  \hline
  \un & 1.03x & 55.14x &  1.11x\\
  \ete & 1.05x & 104.52x & 1.15x\\
  \iwal & 1.05x & 2.34x & 1.07x\\  
  \smd & 1.00x & 12.97x &  1.05x\\
  \pst & 1.03x & 69.31x & 1.12x\\
  \ent & 1.00x & 1.05x & 1.00x\\  
  \un(\ent) & 1.03x & 72.92x & 1.13x\\  
\hline
  \multicolumn{4}{|c|}{\bf Iterative} \\
\hline
  Method & AUCLOG(F1) & Avg(Q's Saved) & Avg(F1) \\
  \hline
  \un & 1.01x & 6.99x & 1.03x \\
  \ete &   1.01x & 6.95x & 1.03x \\
  \iwal & 1.01x & 1.53x & 1.01x\\  
  \smd &  1.01x &  1.47x & 1.00x \\
  \pst &  1.01x & 4.19x & 1.03x  \\
  \ent & 1.01x & 1.46x & 1.01x\\
  \un(\ent) & 1.01x & 1.48x & 1.00x\\  
\hline
\end{tabular}
\caption{{\small Average improvement in AUCLOG, Questions Saved, and Average F1, across  all 15 UCI datasets, by different AL algorithms.}}
\label{tab:avg-improvement}
\vspace{0.1in}
\end{table}

\if{0}
\begin{table*}[t]
\vspace{-.2in}
\small
\centering
\hspace*{-.3in} \begin{tabular}{|m{1.6cm}||m{1cm}|m{1cm}|m{1cm}|m{0.9cm}|c|m{1cm}|m{1cm}|m{1cm}|m{0.9cm}|c|m{1cm}|m{1cm}|m{1cm}|m{0.9cm}|}
 \hline
& \multicolumn{14}{c|}{Ratio of different algorithms to the baseline}\\
 \hline
 & \multicolumn{4}{c|}{AUCLOG of F1 measure} && \multicolumn{4}{c|}{Avg. \# of Questions Saved}  &&  \multicolumn{4}{c|}{Avg.
 F1-measure} \\
 \hline
Dataset & Margin Distance & Bootstrap-LV & Un\-certain\-ty & MinExp Error && Margin Distance & Bootstrap-LV & Un\-certainty & MinExp Error && Margin Distance & Bootstrap-LV & Un\-certainty & MinExp Error\\
\hline
cancer & 1.05x & 1.05x & 1.04x & 1.06x && 14.83x & 17.61x & 15.41x & 25.57x && 1.12x & 1.09x & 1.08x & 1.10x\\
ionosphere & 0.94x & 0.98x & 1.01x & 0.99x && 38.39x & 3.68x & 2.62x & 2.75x && 0.89x & 1.02x & 1.04x & 1.08x\\
optdigits & 1.01x & 1.04x & 1.03x & 1.08x && 11.02x & 48.10x & 42.61x & 136.03x && 1.05x & 1.12x & 1.12x & 1.20x\\
segmentation & 0.85x & 1.07x & 0.97x & 1.10x && 8.57x & 1.17x & 0.72x & 1.25x && 0.87x & 1.39x & 0.96x & 1.51x\\
transfusion & 1.05x & 0.96x & 0.97x & 0.98x && 7.44x & 4.58x & 2.00x & 5.29x && 1.08x & 0.90x & 0.90x & 0.94x\\
yeast & 1.02x & 1.04x & 1.04x & 1.06x && 14.98x & 38.09x & 37.40x & 84.32x && 1.05x & 1.10x & 1.11x & 1.13x\\
glass & 1.03x & 1.12x & 1.19x & 1.15x && 1.33x & 13.19x & 21.67x & 1.74x && 1.30x & 1.59x & 1.86x & 1.62x\\
iris & 1.04x & 1.08x & 1.08x & 1.06x && 4.52x & 190.83x & 27.46x & 63.17x && 1.07x & 1.14x & 1.13x & 1.13x\\
spambase & 1.02x & 1.06x & 1.06x & 1.08x && 13.63x & 570.90x & 556.05x & 1162.79x && 1.05x & 1.17x & 1.16x & 1.19x\\
vehicle & 1.03x & 1.03x & 1.01x & 1.04x && 3.25x & 56.17x & 27.48x & 14.15x && 1.03x & 1.07x & 1.05x & 1.07x\\
haberman & 1.01x & 1.03x & 1.02x & 1.03x && 1.60x & 48.03x & 18.39x & 38.68x && 1.01x & 1.06x & 1.05x & 1.05x\\
mammographic masses & 1.03x & 1.03x & 1.04x & 1.05x && 34.16x & 35.45x & 52.33x & 7.77x && 1.08x & 1.08x & 1.09x & 1.08x\\
pima indians diabetes & 1.01x & 0.99x & 0.99x & 1.03x && 2.04x & 1.97x & 0.77x & 3.00x && 1.03x & 0.96x & 0.96x & 1.08x\\
steel plates faults & 0.87x & 1.02x & 1.05x & 1.04x && 32.68x & 5.85x & 17.79x & 14.16x && 1.02x & 1.03x & 1.08x & 1.05x\\
vertebral column & 1.03x & 1.03x & 1.01x & 1.02x && 6.09x & 3.99x & 4.42x & 7.21x && 1.08x & 1.08x & 1.02x & 1.05x\\
\hline
AVERAGE & 1.00x & 1.03x & 1.03x & 1.05x && 12.97x & 69.31x & 55.14x & 104.52x && 1.05x & 1.12x & 1.11x & 1.15x\\
\hline
\end{tabular}
\caption{Improvement of different AL algorithms over the baseline for the upfront scenario. \srm{Color ``winners'' in each column / row in green}}
\label{tab:uci-upfront}
\end{table*}

\if{0}
\begin{table}[htb]
\small
\centering
\begin{tabular}{|c|c|}
\hline
\centering
{AUCLOG of F1} &  \\
\hline
 Margin Distance & 1.01 \\
Bootstrap-LV & 1.01 \\
Un\-certain\-ty & 1.01 \\
MinExp Error & 1.01 \\
\hline
Avg. Questions Saved & \\
\hline
 Margin Distance & 1.47 \\
Bootstrap-LV & 4.19 \\
Un\-certain\-ty & 6.99 \\
MinExp Error & 6.95 \\
\hline
Avg. F1 Measure  &\\
\hline
 Margin Distance & 1.00 \\
Bootstrap-LV & 1.03 \\
Un\-certain\-ty & 1.03 \\
MinExp Error & 1.03  \\
\hline
\end{tabular}

\caption{Average improvement of different AL algorithms over the baseline for the iterative scenario.}
\label{tab:uci-iterative}
\vspace{-.1in}
\end{table}
\fi

\begin{table*}[htb]
\small
\centering
\hspace*{-.3in} \begin{tabular}{|m{1.6cm}||m{1cm}|m{1cm}|m{1cm}|m{0.9cm}|c|m{1cm}|m{1cm}|m{1cm}|m{0.9cm}|c|m{1cm}|m{1cm}|m{1cm}|m{0.9cm}|}
 \hline
& \multicolumn{14}{c|}{Ratio of different algorithms to the baseline}\\
 \hline
 & \multicolumn{4}{c|}{AUCLOG of F1 measure} && \multicolumn{4}{c|}{Avg. \# of Questions Saved}  &&  \multicolumn{4}{c|}{Avg.
 F1-measure} \\
 \hline
Dataset & Margin Distance & Bootstrap-LV & Un\-certain\-ty & MinExp Error && Margin Distance & Bootstrap-LV & Un\-certainty & MinExp Error && Margin Distance & Bootstrap-LV & Un\-certainty & MinExp Error\\
\hline
cancer & 1.01x & 1.03x & 1.03x & 1.03x && 1.74x & 3.42x & 3.28x & 2.86x && 1.02x & 1.05x & 1.05x & 1.05x\\
ionosphere & 0.99x & 1.00x & 0.98x & 0.94x && 1.05x & 1.15x & 1.41x & 1.34x && 0.99x & 1.01x & 1.03x & 1.02x\\
optdigits & 0.99x & 1.02x & 1.01x & 1.01x && 1.15x & 1.94x & 1.93x & 1.88x && 1.01x & 1.05x & 1.05x & 1.05x\\
segmentation & 1.03x & 1.01x & 1.05x & 1.03x && 1.00x & 1.12x & 1.61x & 1.31x && 1.01x & 1.03x & 1.08x & 1.04x\\
transfusion & 0.97x & 0.85x & 0.88x & 0.82x && 1.18x & 3.53x & N/A & N/A && 0.97x & 0.62x & 0.63x & 0.59x\\
yeast & 1.00x & 1.02x & 1.02x & 1.02x && 1.08x & 3.30x & 3.20x & 3.34x && 1.00x & 1.07x & 1.07x & 1.08x\\
glass & 1.05x & 1.08x & 1.09x & 1.11x && 0.79x & 1.44x & 2.36x & 2.27x && 0.97x & 1.04x & 1.11x & 1.10x\\
iris & 1.01x & 1.01x & 1.01x & 1.02x && 1.76x & 1.76x & 2.58x & 2.59x && 1.00x & 1.00x & 1.00x & 1.00x\\
spambase & 1.00x & 1.03x & 1.03x & 1.03x && 0.98x & 2.84x & 2.82x & 2.95x && 1.01x & 1.07x & 1.07x & 1.08x\\
vehicle & 1.00x & 1.09x & 1.06x & 1.06x && 1.08x & 2.94x & 2.26x & 2.36x && 1.01x & 1.13x & 1.12x & 1.12x\\
haberman & 0.99x & 1.03x & 1.03x & 1.03x && 1.16x & N/A & 0.65x & 1.03x && 0.99x & 1.10x & 1.10x & 1.11x\\
mammographic masses & 1.01x & 1.04x & 1.05x & 1.05x && 4.38x & 28.56x & 69.17x & 68.96x && 1.00x & 1.12x & 1.12x & 1.13x\\
pima indians diabetes & 1.04x & 1.03x & 1.04x & 1.04x && 1.97x & 2.00x & 2.72x & 3.11x && 1.06x & 1.08x & 1.04x & 1.07x\\
steel plates faults & 0.99x & 0.92x & 0.88x & 0.89x && 1.72x & 1.33x & 1.27x & 0.76x && 0.99x & 0.98x & 0.90x & 0.86x\\
vertebral column & 0.99x & 1.03x & 1.02x & 1.03x && 1.06x & 3.33x & 2.68x & 2.55x && 1.00x & 1.08x & 1.07x & 1.08x\\
\hline
AVERAGE & 1.01x & 1.01x & 1.01x & 1.01x && 1.47x & 4.19x & 6.99x & 6.95x && 1.00x & 1.03x & 1.03x & 1.03x\\
\hline
\end{tabular}
\caption{Improvement of different AL algorithms over the baseline for the iterative scenario. \srm{Color ``winners'' in each column / row in green}}
\label{tab:uci-iterative}
\end{table*}
\fi

\eat{
\subsection{Probabilistic Classifiers: Comparison with Bootstrap-LV}
\label{sec:prob}

\begin{table}[h]
\centering
 \begin{tabular}{|l|c|c|}
 \hline
\multicolumn{3}{|c|}{Ratio of \ete to Bootstrap-LV for:}\\
 \hline
Dataset & AUCLOG of F$1$ & Avg. F$1$\\
\hline
cancer & 0.99 & 0.99\\
ionosphere & 1.04 & 1.07\\
optdigits & 0.96 & 0.91\\
segmentation & 1.21 & 1.25\\
transfusion & 0.96 & 0.96\\
yeast & 1.01 & 1.00\\
glass & 1.09 & 1.15\\
iris & 1.04 & 1.07\\
spambase & 1.00 & 0.98\\
vehicle & 1.10 & 1.14\\
haberman & 1.00 & 1.00\\
mammographic masses & 1.02 & 1.04\\
pima indians diabetes & 1.02 & 1.04\\
steel plates faults & 1.69 & 2.36\\
vertebral column & 1.00 & 1.00\\
\hline
AVERAGE & 1.08 & 1.13\\
 \hline

\end{tabular}
\caption{Improvement of \ete over Bootstrap-LV for a probabilistic classifier (i.e., Na\"ive Bayesian Classifier) in the upfront scenario.}
\vspace{-.2in}
\end{table}
}


\subsection{Run-time and Scalability}
\label{sec:exp:scalability}

To measure algorithm runtime, we experimented with multiple datasets.
Here, we only
report the results for the vehicle dataset. 
Figure~\ref{fig:batchsize-runtime} shows that training runtimes
range from about $5,000$ seconds to a few seconds
and
depend heavily on batch size, 
which determines how many times the model is re-trained.

\if{0}

\begin{figure}[th]
\centering
\includegraphics[height=4.8cm]{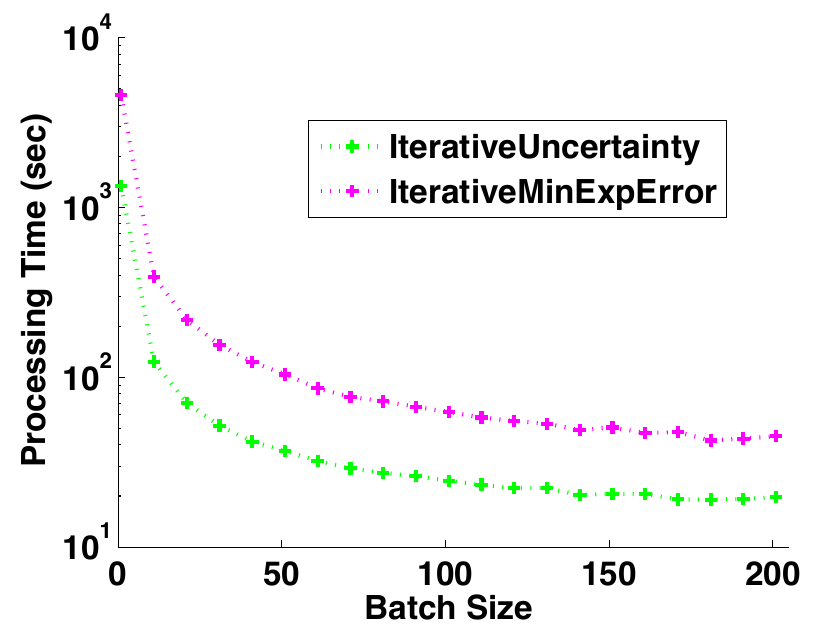}
\caption{The effect of batch size on the processing time of our active learners.}
	\label{fig:batch-time}
\end{figure}

\begin{figure}[th]
\centering
\includegraphics[height=4.8cm]{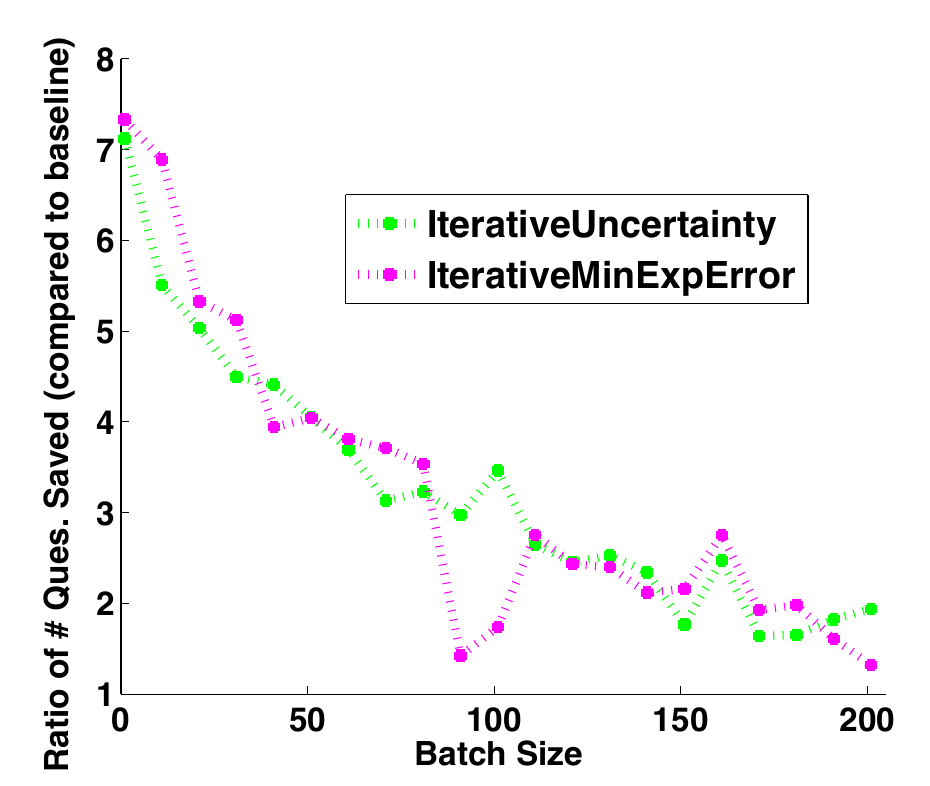}
\caption{The effect of batch size on the quality of our active learners, i.e. the ratio of questions asked by the baseline to those asked by the active learner when achieving the same F1-measure.}
	\label{fig:batch-time}
\end{figure}
\fi

We also studied the effect of parallelism on our algorithms'
runtimes. Here, we compared different AL algorithms in the
upfront scenario on Twitter dataset ($10K$ tweets) as we enabled cores on a
 multicore machine. The results
are shown in Figure~\ref{fig:scale}.  Here, for \un, the run-time only
improves until we have as many cores as we have bootstrap replicas (here, 10). After that,  improvement is marginal.  In contrast, \ete scales extremely well, achieving nearly linear
speedup because it re-runs the model once for every training point.

\begin{figure}[t]
\centering
\includegraphics[height=1.8in, width=1.6in]{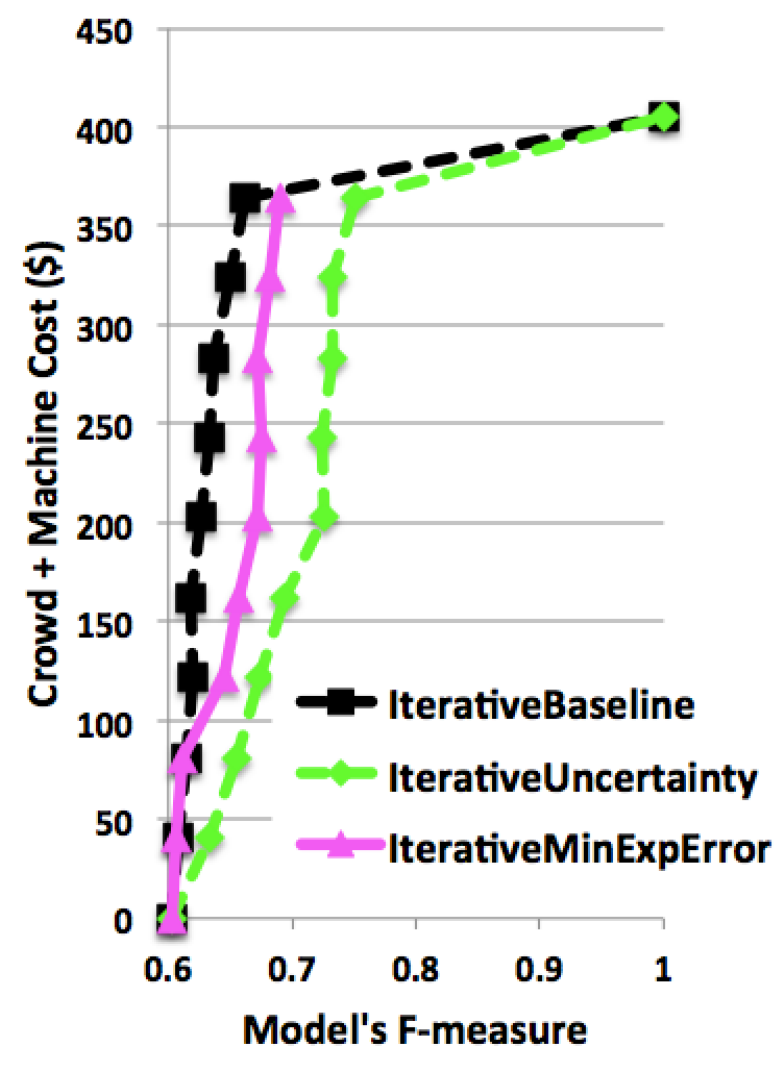}
\caption{{\small The combined monetary cost of crowd and machines for Twitter dataset.}}
\label{fig:monetary-tweet-fmeasure}
\end{figure}

\begin{figure}[t]
\centering
\includegraphics[height=1.8in, width=1.8in]{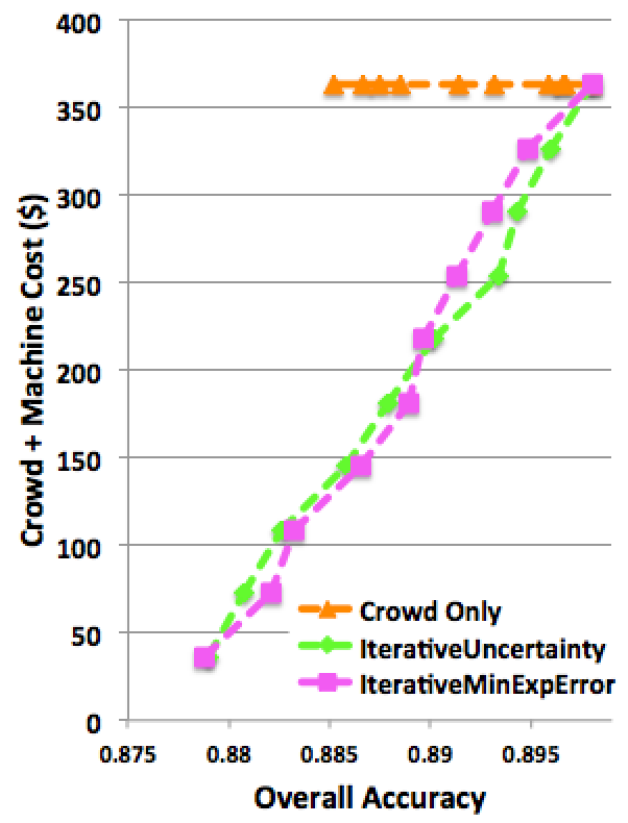}
\caption{{\small The monetary cost of AL vs. using crowd for all labels.}}
\label{fig:monetary-entity-accuracy}
\end{figure}

Finally, we perform a monetary comparison between our AL algorithms and two different baselines.
Figure \ref{fig:monetary-tweet-fmeasure} shows the combined monetary cost (crowd+machines) of 
achieving different levels of quality (i.e., the Model's F1-measure for Twitter dataset from Section \ref{sec:twitter}).
The crowd cost is ($0.01$+$0.005$)*$3$ per labeled item, which includes $3\times$ redundancy and Amazon's commission.
The machine cost for the baseline (passive learner) only consists of  training a classifier while for our  algorithms
we have also included the computation of the AL scores. To compute the machine cost, 
we measured the 
running time in  core-hours using c3.8xlarge instances of Amazon EC2 cloud, which is currently $\$1.68$/hour.
The overall cost is clearly dominated by crowd cost, which is why our AL learners can achieve the same quality 
with a much lower cost (since they ask much fewer questions to the crowd).
We also compared against a second baseline where all the items are labeled by the crowd (i.e.,
no classifiers). As expected, this `Crowd Only' approach is significantly more expensive than our AL algorithms.
Figure \ref{fig:monetary-entity-accuracy} shows that the crowd can label all the items for $\$363$ with an accuracy of 
$88.1$--$89.8\verb=%=$, 
while we can easily achieve a close accuracy of $\$87.9\verb=%=$ 
with only $\$36$ (for labeling
$807$ items and spending less than $\$0.00014$ on machine computation).
This order of magnitude in saved dollars will only become more dramatic over time,
as we expect 
machine costs to continue dropping  according to Moore's law, while 
 human worker costs will presumably remain the same or even increase.

\section{Related Work}
\label{sec:related}

\noindent \textbf{Crowd-sourced Databases.}
These systems \cite{crowdDB, CrowdForge, cdas, crowd-counting, Qurk, deco, getting-it-all}
use optimization techniques to reduce the number of \emph{unnecessary} questions 
 asked to humans (e.g., number of pair-wise comparisons in a join or sort query). 
 However, 
the crowd must still provide at least as many labels as there are unlabeled items directly requested by the user.
It is simply unfeasible to label millions of items in this fashion. To scale up to large datasets,
we  use machine learning to
 avoid obtaining crowd labels for a significant portion of the data.
 
\vspace{0.1in}
\noindent \textbf{Active Learning.} AL has been a rich literature in  machine learning (see \cite{al-survey}).
However, to the best of our knowledge, no existing AL algorithm
 satisfies \emph{all} of the desiderata required for a practical crowd-sourced system, namely generality, black-box approach,
 batching, parallelism, and label-noise management. 
For example,  many AL  algorithms are designed 
for a specific classifier (e.g. neural networks~\cite{jordan-96} or SVM~\cite{svm-margin})
or 
a specific domain (e.g., entity resolution~\cite{arasu-er,active-sampling,sarawagi-er,crowd-er}, vision~\cite{vision-al}, 
or medical imaging~\cite{medical-al}). 
However, our algorithms work for general classification tasks and do not require any domain knowledge.
The popular \iwal algorithm~\cite{iwal} is generic (except for hinge-loss classifiers such as SVM), but does not support batching or parallelism,
and requires adding new constraints to the classifier's internal loss-minimization step.
In fact, most AL proposals that provide theoretical guarantees (i) are not black-box, as they need to \emph{know and shrink} the classifier's hypothesis space
 at each step, and (ii) do not support batching, as they rely on IID-based analysis. 
 Notable exceptions are \cite{general-agnostic} and its \iwal variant~\cite{iwalbb}; they are black-box but they do not support batching or parallelism. 
\pst~\cite{provost} and
\para~\cite{para-active} support parallelism and batching, but both \cite{para-active, iwalbb}
are based on VC-dimension bounds~\cite{ml-textbook}, which are known to be too loose in practice. They cause the model  to
request many more labels than needed, leading to negligible savings over passive learning (as shown in Section~\ref{sec:expr}).
  \pst~also  uses bootstrap, but unlike our algorithms, it is not general.
 Noisy labelers are handled in~\cite{crowd-error, crowd-counting,crowd-filter, GAL}, 
 but \cite{crowd-counting,crowd-filter,GAL} assume the
 same quality for all labelers and~\cite{crowd-error} assumes that each labeler's quality is the same across all items.
Moreover, in Section \ref{sec:expr}, we empirically showed that our algorithms are superior to generic AL algorithms 
(\iwal+\para~\cite{para-active,iwal,iwalbb} and \pst~\cite{provost}).

AL has been applied to many specific domains (some using crowd-sourcing): machine translation \cite{al-cs-1},
 entity resolution \cite{er2, er1, er3, er4, crowd-er, active-sampling},  and SVM classification \cite{svm-margin}. 
Also, \cite{al-cs-2} assumes a probabilistic classifier and \cite{al-sentiment} assumes
access to a clustering algorithm for selecting subsets of articles.
 Our algorithms can handle arbitrary classifiers and do not make any of these assumptions.
Also, surprisingly, we are still 
competitive with (and sometimes even superior to) some of these domain-specific algorithms (e.g., we compared against \smd~\cite{svm-margin}, \crowdER~\cite{crowd-er}, \aditya~\cite{active-sampling}, and Brew et al. \cite{al-sentiment}).

\vspace{0.1in}
\noindent \textbf{Semi-supervised Learning.} AL and semi-supervised learning (SSL) \cite{semi5, semi1,semi2} are closely related. 
SSL  exploits  
the latent structure of unlabeled data.
E.g., \cite{semi1} combines labeled and unlabeled examples 
to infer more accurate 
 labels from the crowd. 
 To achieve  generality, our \pba algorithm 
does not assume 
any prior knowledge of the unlabeled data. 
Another common SSL technique is to train   multiple (ensemble) models  with the labeled data to
 classify the unlabeled data independently, and then use each model's most confident predictions 
  to train the rest \cite{semi5,semi2}. This is similar to our \un algorithm, but we use bootstrap,
    providing
  a generic way for obtaining multiple predictions and unbiased estimates of uncertainty. Also, we
  treat the classifier as a black-box, whereas some of these methods do not.  
However, SSL and AL can be complementary \cite{al-ssl}, and 
thus combining them can be an interesting future work to further improve the scalability of 
crowd-sourced systems. 

\if{0}
Yan et. al~\cite{yan-icml2011} also looked at the problem of AL from a group of workers, but focused on the problem of picking the best worker to answer each question.  This is different from our scenario because in crowd-sourcing systems like Mechanical Turk, the crowd database has no control over which users answer a given item. Pujara et al.~\cite{icml-2011} also propose using AL in a crowd-sourced setting where they ask crowd workers to label the lowest confidence data items.  Additionally, they simply use proximity to the decision boundary as a metric of confidence, which, as we showed in our experiments, does not perform much better than a random baseline. 
\fi

\if{0}
\noindent \textbf{Batch-mode AL.} Batching has been used as a viable approach in AL with large corpus of data~\cite{text-batch}. This is in the same spirit as our iterative scenario.
The problem of batch-size selection in the AL community has been studied (e.g., Guo et al.~\cite{disc-batch}).
 Adapting these algorithms to the crowd-sourced setting described in Section 4.3 would be an interesting direction for future work.
Also, it has been shown that diversifying the items in each batch can improve the results~\cite{batch-diversity}. Our algorithms implicitly diversify each batch through weighted sampling, which ensures that even low-score items have a chance to be labeled (as opposed to a top-K selection strategy). 
\fi

\vspace*{-0.1in}
\section{Conclusions}
\label{sec:conclusion}

In this paper, we proposed two AL algorithms, \un and \ete, to enable 
 crowd-sourced databases to scale up to large datasets.
To broaden their applicability to different classification tasks, we designed these algorithms based on the theory of nonparametric bootstrap and evaluated them in two different settings.
In the {\it
  upfront} setting, we ask all questions to the crowd in one go.  
  In the {\it
  iterative} setting, the questions are adaptively picked and added to the labeled pool. Then, we retrain the model
and repeat this process. While iterative retraining is more expensive, it also has a higher chance of learning a better model. 
Additionally, we proposed algorithms for
choosing the number of questions to ask different crowd-workers, based on the characteristics of the data being
labeled. We also studied the effect of batching on the overall runtime and
quality of our AL algorithms. Our results, on three data
sets collected with Amazon's Mechanical Turk, and with $15$ datasets
from the UCI KDD archive, show that our algorithms make substantially fewer label requests than state-of-the-art AL techniques.
  We believe that these algorithms would
prove to be immensely useful in crowd-sourced database systems. 

\vspace*{-0.1in}


{\small
\bibliographystyle{abbrv}
\bibliography{learning,mozafari}

\end{document}